\documentclass[nonacm]{jair}

\AtBeginDocument{%
\fancyfoot{}%
\fancypagestyle{firstpagestyle}{\fancyhf{}\fancyfoot{}}%
}

\usepackage{fix-cm}

\usepackage{tcolorbox}
\usepackage{latexsym}
\usepackage{booktabs}
\usepackage{enumitem}
\usepackage{graphicx}
\usepackage{color}

\usepackage{multirow} 
\usepackage{algorithm}
\usepackage{algorithmic}
\usepackage{array}
\usepackage{listings}
\newfloat{listing}{tb}{lst}{}
\floatname{listing}{Listing}

\setcopyright{cc}
\copyrightyear{2025}
\acmDOI{10.1613/jair.1.xxxxx}

\JAIRAE{Insert JAIR AE Name}
\JAIRTrack{} 
\acmVolume{4}
\acmArticle{6}
\acmMonth{8}
\acmYear{2025}

\RequirePackage[
  datamodel=acmdatamodel,
  style=acmauthoryear,
  backend=biber,
  giveninits=true,
  uniquename=init
  ]{biblatex}

\begin{document}

\title{DCP-Bench-Open: Evaluating LLMs for Constraint Modelling of Discrete Combinatorial Problems}

\author{Kostis Michailidis}
\authornote{Corresponding Author.}
\orcid{0009-0000-2139-0106}
\email{kostis.michailidis@kuleuven.be}
\affiliation{%
  \institution{KU Leuven}
  \country{Belgium}
}

\author{Dimos Tsouros}
\orcid{0000-0002-3040-0959}
\email{dtsouros@uowm.gr}
\affiliation{%
  \institution{University of Western Macedonia}
  \country{Greece}}

\author{Tias Guns}
\orcid{0000-0002-2156-2155}
\email{tias.guns@kuleuven.be}
\affiliation{%
  \institution{KU Leuven}
  \country{Belgium}
}


\renewcommand{\shortauthors}{Michailidis, Tsouros \& Guns}


\begin{abstract}
Discrete Combinatorial Problems (DCPs) are prevalent in industrial decision-making and optimisation. However, while constraint solving technologies for DCPs have advanced significantly, the core process of formalising them—namely constraint modelling—requires significant expertise and remains a bottleneck for wider adoption. Aiming to alleviate this bottleneck, recent studies have explored using Large Language Models (LLMs) to transform combinatorial problem descriptions into executable constraint models. However, the existing evaluation datasets for discrete constraint modelling are often limited to small, homogeneous, or domain-specific problems, which do not capture the diversity of real-world scenarios. This work addresses this gap by introducing \textbf{DCP-Bench-Open}, a novel benchmark that includes a diverse set of well-known discrete combinatorial problems sourced from the Constraint Programming (CP) and Operations Research (OR) communities, structured explicitly for evaluating LLM-driven constraint modelling. With this dataset, and given the variety of modelling frameworks, we compare and evaluate the modelling capabilities of LLMs for three distinct constraint modelling systems, which vary in abstraction level and underlying syntax. Notably, the results show higher performance when modelling with a high-level Python-based framework. Additionally, we systematically evaluate the use of prompt-based and inference-time compute methods across different LLMs, which further increase accuracy, reaching up to 91\% on this highly challenging benchmark. DCP-Bench-Open is publicly available in \url{https://github.com/DCP-Bench/DCP-Bench-Open}.
\end{abstract}

\maketitle

\section{Introduction}
\label{sec:introduction}

Discrete combinatorial problems exist in numerous real-world applications, where optimal decision-making is essential, including logistics, scheduling, and network design \citep{paschos2014applications}.
Solving these problems requires finding the best combination of elements under specific constraints, which is computationally intensive; many such problems are NP-hard, meaning that their solving time can increase exponentially with problem size in the worst case.
This challenge has led to the development of various industrial solvers and solving paradigms, including Constraint Programming (CP), Integer Linear Programming (ILP) and Boolean Satisfiability (SAT),  among others \citep{paschos2014applications,wallace1996practical,simonis1999building}.

Common to most solving paradigms is the need for an input formal model: a declarative specification in which users specify \textit{what} constraints the solution must satisfy, rather than detailing \textit{how} to find it.
Despite the effectiveness of such solving approaches, the complex process of modelling—translating a problem description into a formal specification—remains a significant bottleneck.
Specifically for CP, modelling involves identifying and formally defining the problem's decision variables with their domains, constraints on the variables, and, where applicable, an objective function.
This process demands a deep understanding of the application domain and proficiency in the chosen modelling framework, both syntactically and semantically.
This modelling complexity is recognised as one of the most important limitations for the wider use of combinatorial optimisation techniques, restricting accessibility for non-experts \citep{freuder2014grand,freuder2018progress}.

\begin{figure}[t]
    \includegraphics[width=0.75\linewidth]{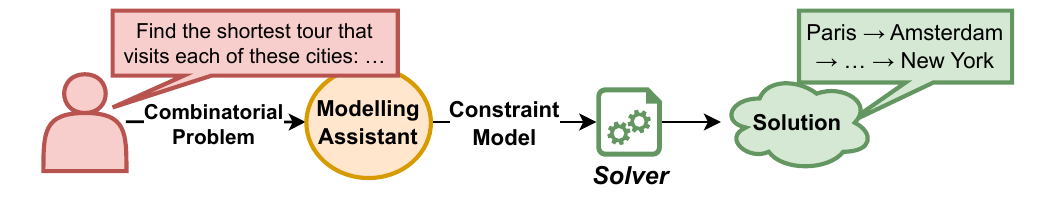}
    \Description{A flowchart illustrating a process where a user provides a combinatorial problem in natural language, such as "Find the shortest tour that visits each of these cities...". This input is processed by a "Modelling Assistant," which converts it into a "Constraint Model." The constraint model is then sent to a "Solver," which produces a final "Solution," depicted as a route "Paris -> Amsterdam -> ... -> New York."}
    \caption{LLM-driven constraint modelling: users state a combinatorial problem in natural language, which the system transforms into a formal constraint representation, and delegates the latter to a constraint solver.}
    \label{fig:pipeline_io}
\end{figure}

This raises a central question: Can we lower the modelling barrier to make CP and related technologies for Discrete Combinatorial Problems accessible to a broader audience?
Prior works in this direction include constraint acquisition methods that aim to learn the constraints from data \cite{bessiere2007query,bessiere2023learning,mechqrane2024using} and constraint detection methods from natural language \cite{kiziltan_detection_2016}.
More recently, advances in large language models (LLMs) offer the opportunity of creating \textit{modelling assistants}~\cite{cappart2025combining} that could initiate or (semi-)automate the modelling process (as illustrated in Figure \ref{fig:pipeline_io}), in analogy to how coding assistants are widely being used for a variety of coding tasks \cite{jiang2024codeLLMs,chen2021evaluating,fan2023automated}.

Nevertheless, LLM-driven constraint modelling presents different challenges compared to general code generation.
First, while imperative code specifies step-by-step sequential instructions, declarative modelling requires that all constraints must hold true simultaneously.
Second, formalizing the extracted conceptual constraints is hard due to the combinatorial nature of modelling choices themselves. Specifically, modelling requires selecting a \textit{viewpoint}—a specific way to represent variables and structure constraints based on the problem domain—from potentially many alternatives \cite{frisch2005rules}.
Third, verifying the correctness of a generated constraint model is different from imperative code generation, in which correctness can often be tested using example inputs and expected outputs (e.g., unit tests). In contrast, constraint modelling lacks such verifications, as a “test” would require knowing the correct model or solution in advance.

Recent works have explored the potential of LLMs in this direction by generating solvable representations of numerical reasoning problems in formal declarative modelling frameworks \citep{ye_satlm_2023,tsouros_holy_2023}. \citet{ishay_leveraging_2023} frame Logic Grid Puzzles (LGPs) as Answer Set Programming (ASP) programs, \citet{prasath_synthesis_2023} focus on linear optimization problems, and \citet{ye_satlm_2023} propose translating inputs into satisfiability problems. Although promising results have been achieved for simple domain-specific problems, such as small-scale linear optimization problems \citep{ramamonjison_nl4opt_2023,prasath_synthesis_2023} and LGPs \citep{ishay_leveraging_2023}, scaling to more complex and diverse problems remains a significant challenge \citep{michailidis2024constraint}.

Existing limitations can be summarized as follows. (a) Current NL-to-CP and other benchmarks lack the diversity and complexity necessary to represent a wide range of realistic discrete combinatorial problems. This constitutes a prerequisite for assessing LLM capabilities on this task. (b) Given the wide variety of constraint modelling frameworks available, ranging in abstraction levels and interface types, there is a need for systematic evaluation across them to explore their impact on LLM-driven model generation. (c) Applying recent LLM advances to combinatorial modelling remains underexplored, even though test-time scaling methods have shown success for complex programming tasks \citep{chen2023teaching}.

To address the lack of representative benchmarks, we developed DCP-Bench-Open (extending the previous CP-Bench~\citep{michailidis2025cp}), a novel dataset of discrete combinatorial problems gathered from distinct sources within the research community.
While existing repositories (e.g., PyCSP3 examples \cite{lecoutre2020pycsp3}, CSPLib \cite{csplib}, and others) contain constraint models, they are not suitable for automated evaluation of LLM-generated models, as they are not structured for this purpose.
We designed DCP-Bench-Open for this task based on solution-level evaluation, which is non-trivial since more than one (optimal) solution may exist for a combinatorial problem (detailed in Section \ref{subsec:evaluation}).

Using this benchmark, we systematically evaluate the capabilities of state-of-the-art LLMs when prompted to generate models using different modeling frameworks, varying by abstraction level (low vs. high) and interface type (domain-specific vs. Python-based).
We select the following as representatives: OR-Tools \citep{ortools,cpsatlp} offers a low-level, direct solver Python API; CPMpy \citep{guns2019increasing} provides a high-level, Python-based modelling interface; and MiniZinc \citep{minizinc} is a well-established high-level, domain-specific modelling language.
Across these frameworks, we compare different system prompts, starting from basic minimal instructions and progressing to detailed, classroom-level, modeling guidelines and including documentation specific to the framework used.

With inference-time computation showing high success rates across various LLM tasks \cite{snell2024scaling}, we adapt and evaluate such methods in order to further enhance the quality of the generated constraint models.
Firstly, we use Retrieval-Augmented In-Context Learning (RAICL) to enrich the prompt context with representative examples \citep{ye-etal-2023-complementary,michailidis2024constraint}.
Secondly, we investigate reasoning LLMs~\cite{reasoning_survey} that utilize internal chain-of-thought processes to decompose the modelling procedure.
Thidly, we exploit the probabilistic nature of LLMs, by repeatedly sampling multiple CP models and employing solution majority voting to select the most reliable model based on the observed outputs \citep{chen2021evaluating,olausson2023linc}.
Finally, to address detectable errors, such as runtime or solution printing ones, and trigger self-correction of modelling errors, we investigate the effect of self-verification prompting on LLMs for iterative model refinement.

The primary contributions of this work are the following:
\begin{itemize}
    \item A novel benchmark structured specifically for evaluating LLM-driven constraint modelling of discrete combinatorial multi-instance problems. The dataset consists of 164 diverse combinatorial problem descriptions along with their runnable constraint models, 23 of which contain more than one data instance.
    \item Systematic evaluation of state-of-the-art LLMs across three commonly used modelling frameworks, showing the impact of interface type on accuracy, with Python-based frameworks achieving up to 75\% accuracy compared to a maximum of 57.3\% for the domain-specific language in zero-shot settings.
    \item Adaptation of prompt-based and inference-time compute methods for enhancing LLM-driven constraint modelling, reaching up to 91\% accuracy.
\end{itemize}

\paragraph{Publication History} This version extends the work presented in \citet{michailidis2025cp} in three central directions. Firstly, we expand the benchmark from 101 to 164 problems, also incorporating problems from the ComplexOR dataset and facilitating the inclusion of multiple data instances per problem. Importantly, as some of the original problem instances contained underspecified descriptions or overconstrained ground-truth models (e.g. due to symmetry-breaking constraints), we also manually verified and corrected them\footnote{More details on the affected instances can be found in the \href{https://huggingface.co/datasets/kostis-init/CP-Bench/blob/main/cp_bench_changes.md}{CP-Bench repository}.} in this new version. Secondly, we introduce a Multi-Instance evaluation framework to test model generalization across hidden data instances, with additional experiments showing significantly lower accuracy on the new strictest metric. Thirdly, we evaluate a new generation of LLMs with a final inference-time compute experiment across all three modelling frameworks, showing large improvements in accuracy of all frameworks. Furthermore, more minor changes in the experiments include: a) adding self-verification in the error types experiment (Q2), and b) the addition of "reasoning" as an additional inference-time compute method in Q3. Finally, we added a dedicated section discussing the Related Work.

\section{Background}
\label{sec:background}

\subsection{Constraint Programming}
\label{subsec:cop}

We use CP as a formal intermediate representation for the given combinatorial problems.
A \textit{Constraint Optimization Problem} (COP) is a type of combinatorial problem that involves finding an optimal assignment of values to decision variables, subject to a set of constraints and an objective function.
Formally, a COP is defined as a tuple \((\mathbf{X}, \mathbf{D}, \mathbf{C}, f)\) where:

\begin{itemize}
    \item \(\mathbf{X} = \{x_1, x_2, \ldots, x_n\}\) is a set of \(n\) decision variables.
    \item \(\mathbf{D} = \{D_1, D_2, \ldots, D_n\}\) is a set of \(n\) domains, where each \(D_i\) is a finite set of allowable values for \(x_i\).
    \item \(\mathbf{C} = \{C_1, C_2, \ldots, C_m\}\) is a set of \(m\) constraints, with each \(C_j\) specifying allowed combinations of values for a subset of variables. Formally, \(C_j \subseteq D_{j1} \times D_{j2} \times \ldots \times D_{jk}\) for some subset \(\{x_{j1}, x_{j2}, \ldots, x_{jk}\} \subseteq \mathbf{X}\).
    \item \(f: \prod_{i=1}^{n} D_i \rightarrow \mathbb{R}\) is the objective function to be optimized (either maximized or minimized).
\end{itemize}

The goal is to find an assignment \(\mathbf{a} = \{x_1 = v_1, x_2 = v_2, \ldots, x_n = v_n\}\) that satisfies all constraints in \(\mathbf{C}\) and optimizes the objective function \(f\). Formally, an assignment \(\mathbf{a}\) is feasible if \(v_i \in D_i\) for all \(i\) and if \((v_{j1}, v_{j2}, \ldots, v_{jk}) \in C_j\) for each constraint \(C_j\). An optimal solution \(\mathbf{a}^*\) is a feasible assignment that optimizes \(f\).
If no objective function $f$ is defined, the problem reduces to a \textit{Constraint Satisfaction Problem} (CSP), where the goal is to find any feasible assignment. When no feasible assignment exists, the problem is considered unsatisfiable.
For brevity, we will use $\mathbf{a}^*$ to represent either an optimal solution in COPs or a feasible solution in CSPs, and we will commonly refer to it as the \textit{optimal solution}.

Notably, CP systems are not restricted to linear constraints but also support a variety of global constraints that capture common patterns in combinatorial problems. These include min/max aggregators, element constraints (for array access), count/nr-values (for aggregation over arrays), cumulative constraints (for scheduling-type problems), and many others that facilitate more expressive modelling of complex problems \citep{beldiceanu2007global}.

\subsection{Large Language Models} 
\label{subsec:large_language_models}

Based on the Transformer architecture \citep{vaswani_attention_2023}, LLMs are deep learning systems, typically with billions of parameters, capable of learning and generating complex language patterns \citep{plm_survey_2021}. In this work, we use them as black-box text generators, configured only by an input text (or prompt) and a randomness parameter (temperature). Formally:

\begin{equation} 
\label{eq:llms}
\begin{split}
LLM(p, \tau) = (x_1, x_2, \ldots, x_w), \quad \text{where } \\
x_{t+1} \sim P_{\tau}(x_{t+1} \mid p, x_{\leq t})
\end{split}
\end{equation}

Here, \( x_{\leq t} \) denotes the sequence of tokens generated up to time step \( t \), with \( x_{\leq 0} \) as the empty sequence and \( x_1 \) as the first generated token. At each step, the next token \( x_{t+1} \) is sampled from the temperature-adjusted probability distribution \( P_{\tau}(x_{t+1} \mid p, x_{\leq t}) \), which is typically computed by applying the softmax function to the model's logits—raw unnormalized scores over the vocabulary. The temperature \( \tau \) scales the logits before softmax, with higher values increasing randomness and diversity, while lower values favour the most likely tokens. The generation process continues until a predefined stop token is encountered or the maximum context length is reached. We assume tokenization and detokenization into text are handled internally by the LLM.

\section{Problem Formulation}
\label{sec:formulation}

As illustrated in Figure~\ref{fig:pipeline_io}, we conceptualize a system capable of assisting users in modelling combinatorial problems, transforming their textual descriptions into executable models that a constraint solver can solve. In this paper, we focus on this transformation, which requires a mechanism that can process and accurately represent the problem's constraints, variables, and objectives—tasks that demand expert knowledge in both the application domain and CP modelling.
We consider LLMs such versatile tools, because of their powerful natural language processing and coding capabilities \citep{chen2021evaluating}.


Formally, we define the process as follows:
Let \( \mathcal{V}^* \) represent the set of all possible text sequences composed of tokens from the LLM's vocabulary \( \mathcal{V} \). Let \( \mathcal{S} \) be the space of all possible valid assignments (solutions) to the decision variables of a discrete combinatorial problem. 
We utilize the previously defined function \( LLM \), which maps an input prompt\footnote{We assume that input \( p \) includes bothP instructions for a modelling framework and a problem description (which includes a default instance of parameter values for parameterized problems).} \( p \in \mathcal{V}^* \) and a temperature parameter \( \tau \in \mathbb{R} \) to an output sequence \( \pi \in \mathcal{V}^* \). Here, \( \pi \) represents an \textit{executable program} (e.g., a Python script) that encodes the constraint model.
Next, we define an execution function  \( exec: \mathcal{V}^* \rightarrow \mathcal{S} \cup \{\emptyset, error\} \), which acts as a runtime environment that interprets the program \( \pi \).
The function yields an output \( \mathbf{a}^+ \), which represents either an optimal solution $ \mathbf{a}^* \in \mathcal{S} $, an unsatisfiable result $\emptyset$, or some kind of error (e.g., syntax, runtime, solver timeout etc.). The overall system can be represented as follows:
\begin{equation}
    exec(LLM(p, \tau)) = \mathbf{a}^+ 
\end{equation}

\section{Methodology}
\label{sec:methodology}

The task for LLMs in this context is to translate problem descriptions into a declarative specification of variables and constraints over them, including expressing them correctly in a specific modelling framework and generating code for printing the computed solution in a structured way. These latter requirements demand coding skills; thus, we take inspiration from successful test-time scaling techniques used for generating imperative code.

\subsection{System Prompt}
\label{subsec:system_prompt}

Given the complexity of translating natural language descriptions into constraint models, we explore different levels of system prompts to guide LLMs more effectively in a 0-shot setting. Without examples in the prompt (hence 0-shot), providing a comprehensive description of the task is important to help the LLM process and correctly model the problem declaratively.
Formally, if \( sys \in \mathcal{V}^* \) is the system prompt describing the task, and \( c \in \mathcal{V}^* \) is the combinatorial problem description, then the LLM input prompt is \( p = sys \oplus c \), concatenating the system prompt and the problem description.
We examine three levels of system prompts, each progressively providing more detailed information (the prompts are available in the supplementary material):

\begin{enumerate} 
    \item \textbf{Basic prompt.} The basic prompt provides essential instructions to the LLM, only describing the task of producing a constraint model using a specific framework and how the solution should be formatted. This prompt simulates a scenario where a user gives minimal guidance, e.g. as one would do in a chat interface.

    \item \textbf{Guidelines prompt.} Expanding on the basic prompt, this level adds specific guidelines related to the modelling process. These guidelines include code generation steps, generic classroom-style advice on constraint modelling, and a template on how to format the response model in the modelling framework used. It mirrors a situation where a user provides more elaborate but still generic instructions.

    \item \textbf{Documentation prompt.} The most detailed prompt builds on top of the guidelines prompt, appending single-line API documentation of the available classes and functions of the CP modelling framework. 
    This prompt offers (limited) access to documentation, enabling direct reference to available methods and functionalities.
\end{enumerate} 

\noindent The Basic prompt that we used for the CPMpy framework is shown in Listing~\ref{lst:system_prompt_basic}. All system prompts for each level and framework are available in detail in Appendix~\ref{sec:app_sys_prompts}.

\begin{listing}[ht]
\caption{Basic System Prompt (Level 1) for the CPMpy framework}
\label{lst:system_prompt_basic}
    \begin{tcolorbox}[colback=gray!10, colframe=gray!60, boxrule=0.5pt, arc=2pt]
        \ttfamily\small
        
        You are an expert in Python, constraint programming, and modelling combinatorial problems.
        Your task is to model the given problem using the CPMpy library.
        Specifically, you should generate Python code that uses CPMpy to define and solve the problem.
        Only the standard Python libraries, CPMpy, and numpy should be used.

        \vspace{0.5em}
        \textbf{\# Output Formatting}

        Here is an example for printing; assume the problem description contains:
        "Print the number of apples and oranges (apples, oranges), the cost per round (cost), and the total cost (total\_cost)."
        
        In this case, the output of the solution as JSON should be done as follows:
        
        \begin{verbatim}
```python
if model.solve():
    # assuming 'apples', 'oranges', 'cost' are variables
    solution = {
        'apples': int(apples.value()), 'oranges': int(oranges.value()),
        'cost': cost.value().tolist(), 'total_cost': int(model.objective_value())
    }
    print(json.dumps(solution))
else:
    print("No solution found.")
```\end{verbatim}
    \end{tcolorbox}
\end{listing}

\subsection{In-Context Learning}
\label{subsec:raicl}

Few-shot prompting is a widely used method that includes task-relevant examples in the input prompt, improving LLM responses without requiring retraining \citep{brown_language_2020}. However, selecting effective examples is critical, as a static or random set of examples may not align well with diverse inputs \citep{liu_what_2022,ye_compositional_2023}.
To address this, recent studies have examined dynamically inserting examples in the prompt context based on the current input \citep{ye-etal-2023-complementary}. In the context of constraint modelling, adding examples of semantically similar problems and their corresponding models has previously improved the accuracy of generated models \citep{michailidis2024constraint}.

\begin{figure}[ht]
    \centering
    \includegraphics[width=0.75\linewidth]{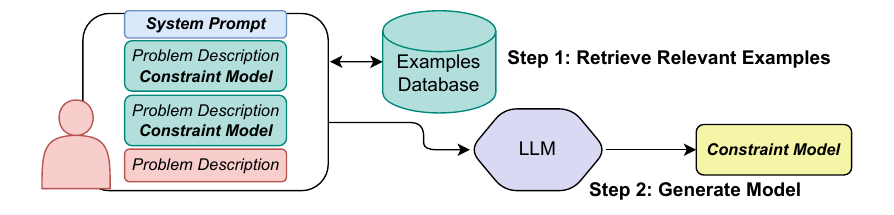}
    \Description{A diagram illustrating a two-step process. In Step 1, relevant examples (green boxes) are retrieved from an "Examples Database" and combined with a "System Prompt" and the user's current "Problem Description" (red box). In Step 2, this constructed prompt is fed into an "LLM" (purple hexagon), which generates the final "Constraint Model" (yellow box).}
    \caption{Retrieval-augmented few-shot prompting: on-the-fly example retrieval to provide more relevant patterns to the LLM.}
    \label{fig:in_context_retrieval}
\end{figure}

Formally, this method retrieves from a prebuilt database an ordered set \( E \) of \( e \) input-output pairs, \( E = \left\{(i_j, o_j)\right\}_{j=1}^e \), based on their relation to the given problem \( c \). Given a database of problem-model pairs, the retrieval process involves comparing the token embeddings of \( c \) with each problem in the database and selecting the \( e \) most similar ones using a predefined metric, such as semantic similarity \cite{semantic_sim}. The prompt then becomes \( p = sys \oplus E \oplus c \), where the retrieved examples \( E \) are placed between the system prompt and the current problem description, as shown in Figure~\ref{fig:in_context_retrieval}.

\subsection{Repeated Sampling}
\label{subsec:repeated_sampling}

While enriching the prompt with instructions and examples can enhance the guidance provided to LLMs, it does not fully address the challenge of synthesizing CP models from textual descriptions.
By generating a broad distribution of potential solutions, we can increase the probability that at least one candidate is correct. In this direction, we aim to benefit from the probabilistic nature of LLMs through repeated sampling (or self-consistency) \citep{chen2021evaluating}. Instead of relying on the single “most probable” output, we increase the temperature parameter value $\tau$, generating multiple candidate models for the same problem description. This strategy exploits the variability in LLM outputs to identify the most consistent model across multiple runs.

Specifically, we apply \textit{Solution Majority Voting} for the final selection of the generated model~\citep{olausson2023linc}. By repeatedly querying the LLM and solving each generated constraint model, we observe and save the solutions they produce; then, the first model yielding the most frequently observed solution is selected.
If none of the models produces a solution, we default to the first generated model.
The process is outlined in Algorithm~\ref{alg:repeated_sampling}.

\begin{algorithm}[h]
\caption{Sampling with Solution Majority Voting}
\label{alg:repeated_sampling}
\begin{algorithmic}[1]
\REQUIRE \( (p, k, \tau) \) \quad // Prompt, \#Samples, Temperature
\STATE \( \mathcal{C}, \mathcal{S} \gets \emptyset, \emptyset \) \quad // Candidate CP models \& solutions
\FOR{\( i = 1 \) to \( k \)}
    \STATE \( m_i \gets LLM(p, \tau) \) \quad // Generate CP model
    \STATE \( s_i \gets Solver(m_i) \) \quad // Extract the solution
    \STATE \( \mathcal{C}, \mathcal{S} \gets \mathcal{C} \cup \{m_i\}, \mathcal{S} \cup \{s_i\} \)
\ENDFOR
\STATE \( s_{\text{maj}} \gets \text{most\_freq}(\mathcal{S}) \) \quad // Find most frequent solution
\IF{\( s_{\text{maj}} \) exists}
    \RETURN \( \mathcal{C}[\text{first\_index\_of}(s_{\text{maj}}, \mathcal{S})] \)
\ENDIF
\RETURN \( \mathcal{C}[1] \) \quad // Fallback to first candidate model
\end{algorithmic}
\end{algorithm}

\subsection{Self-Verification}
\label{subsec:self-debug}

Writing semantically and syntactically correct code in a single pass, without the ability to revise previous steps, is a complex task even for experienced programmers.
This challenge holds even more for CP modelling, where variables need to be declared before constraints, and constraints should not contradict earlier ones.
We expect that the generated models may occasionally contain runtime, modelling or formatting errors. This motivates us to define three criteria that assert the validity of the generated constraint model:

\begin{itemize}
\item \textbf{Runtime Integrity:} The generated code executes without errors and correctly utilizes library components.
\item \textbf{Model Accuracy:} The decision variables, constraints, and objective function are correctly defined and aligned with the problem description; when the model is executed, the produced solution is valid and optimal.
\item \textbf{Solution Output:} The solution is printed in the correct structured format as per the user's provided instructions.
\end{itemize}

\noindent A promising approach to improve the reliability of code generation is to allow LLMs to review, verify, and refine their previous output, effectively mimicking the human debugging process \citep{chen2023teaching,weng-etal-2023-large}.

\begin{figure}[ht]
    \centering
    \includegraphics[width=0.7\linewidth]{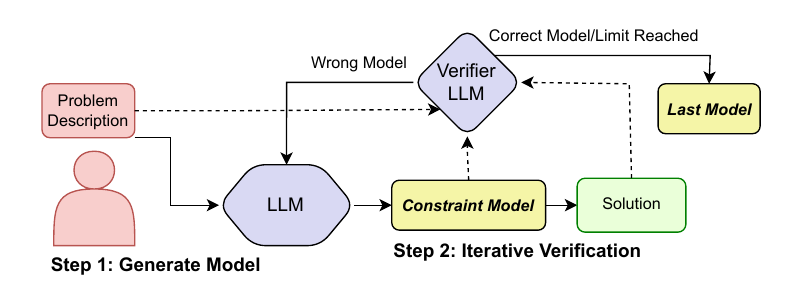}
    \Description{A flowchart depicting "Step 1: Generate Model" and "Step 2: Iterative Verification." A "Problem Description" is input to an "LLM," which produces a "Constraint Model." This model yields a "Solution." The "Verifier LLM" evaluates the "Constraint Model," "Solution," and original "Problem Description." If the verdict is "Wrong Model," feedback loops back to the LLM for regeneration. If "Correct Model/Limit Reached," the process ends at "Last Model."}
    \caption{Iterative Self-Verification: the generated CP model is verified iteratively; both the original problem description and produced solution are also provided back to the LLM. In this work, we use a single LLM for both model generation and verification (thus, self-verification).}
    \label{fig:self_debugging}
\end{figure}

We composed a \textit{Self-Verification} base prompt $v$ (Listing~\ref{lst:self_debug_prompt}) that guides the LLM through an evaluation and correction process, instructing it to iteratively assess the model based on the aforementioned criteria.
Appended to this prompt are: (a) the system prompt $sys$, (b) the problem description $c$, (c) the model $m$ generated by the LLM in the previous attempt, and (d) its resulting output $ \mathbf{a}^+ $ when executed, which may be either a solution of the model or a runtime error traceback message.
Formally, the full self-verification prompt at iteration $t$ is denoted as $v \oplus sys \oplus c \oplus m_{t-1} \oplus \mathbf{a}^+_{t-1} $.
This process continues until either the LLM determines that the most recent constraint model $m_{t-1}$ is correct according to the specified criteria, or a predefined number of iterations is reached; at that point, the latest model is selected. The process is depicted in Figure~\ref{fig:self_debugging}.

\begin{listing}[ht]
\caption{Base Self-Debug Prompt}
\label{lst:self_debug_prompt}
    \begin{tcolorbox}[colback=orange!10, colframe=orange!60, boxrule=0.5pt, arc=2pt]
        \ttfamily\small 

You are an expert in combinatorial optimization and you are asked to verify (or debug) a code snippet that models a combinatorial problem.
You will be given a combinatorial problem description with instructions on how to model it, its code formulation according to these instructions, and the code output.
Explain the given code, especially elaborating on the decision variables, constraints, and the objective function (if applicable).

Then, evaluate the code's correctness in three aspects:
\begin{enumerate}
\item \textbf{Runtime:} Does the code run successfully without syntax errors, and does it correctly utilize the required libraries?
\item \textbf{Model:} Are the decision variables, constraints, and objective function (if applicable) correctly defined? Does the generated solution satisfy the constraints and objective of the given problem description?
\item \textbf{Solution Printing:} Does the code print the solution in the required JSON format, with the correct keys and values according to the given instructions?
\end{enumerate}
If the code is correct, end your response with [[OK]].
If the code is incorrect, provide a corrected version of the code between triple backticks, ensuring the fixed code is self-contained and runnable. End your response with [[FIXED]].

Note: Use [[OK]] and [[FIXED]] only once at the end of your response, and only one of them.
    \end{tcolorbox}
\end{listing}

\section{DCP-Bench-Open}
\label{sec:dataset}

The evaluation of LLMs on translating textual problem descriptions into formal CP specifications has so far been limited by the scope and diversity of available datasets. Existing benchmarks, such as NL4Opt \citep{ramamonjison_augmenting_2022}--part of a competition at NeurIPS\footnote{\url{https://neurips.cc/virtual/2022/competition/50079}}-- and \textit{Logic Grid Puzzles} (LGPs) \citep{jabrayilzade_lgpsolver_2020}, primarily feature small, domain-specific problems, such as simple linear programming (LP) problems or homogeneous puzzles, as summarized in Table~\ref{tab:dataset_comparison}.
More recently, \citet{singirikonda2025text2zinc} presented a cross-domain dataset with MiniZinc models, focusing on CP, LP and Mixed-Integer Programming (MIP) problems as well.

\begin{table}[ht]
    \caption{Comparison of DCP-Bench-Open v0.1.0 with existing datasets for evaluating LLM-generated constraint models of combinatorial problems. Unique constraints count distinct constraint operators and their argument structures (arity and type, e.g., variable/constant/expression). The number of problems reflects the test subset of each dataset. The drop in some average values is due to the changes and removal of symmtery-breaking constraints during the verification and correction of the original (CP-Bench) ground-truth models towards this new version.}    
    \centering
    \small
    \resizebox{\textwidth}{!}{%
    \begin{tabular}{llcccccc}
        \toprule
        \textbf{Benchmark} & \textbf{\#} & \multicolumn{2}{c}{\textbf{Constraints}} & \multicolumn{2}{c}{\textbf{Decision Variables}} & \textbf{\# Unique} & \textbf{\# Opt.} \\
        & & \textbf{Mean (Median, IQR)} & \textbf{Min/Max} & \textbf{Mean (Median, IQR)} & \textbf{Min/Max} & \textbf{Constr.} & \textbf{Probs.} \\
        \midrule
        \textbf{NL4Opt} & 289 & 2.90 (3.0, 1.0) & 2/5 & 2.02 (2.0, 0.0) & 2/3 & 27 & All \\
        \textbf{LGPs} & 100 & 38.47 (38.0, 29.0) & 7/69 & 12.00 (12.0, 0.0) & 12/12 & 8 & None \\
        \midrule
        \textbf{CP-Bench} \textit{[old]} & 101 & 84.76 (13.5, 45.0) & 1/2017 & 51.69 (13.0, 17.3) & 3/962 & 241 & 30 \\
        \textbf{DCP-Bench-Open} \textit{[new]} & 164 & 69.48 (12.0, 33.5) & 1/2463 & 40.33 (13.0, 25.5) & 2/716 & 349 & 54 \\
        \bottomrule
    \end{tabular}%
    }
    \label{tab:dataset_comparison}
\end{table}

To address the gap in discrete CP-modelling LLM benchmarks~\cite{michailidis2024constraint}, we introduce \textbf{DCP-Bench-Open}\footnote{\url{https://github.com/DCP-Bench/DCP-Bench-Open}.} \cite{dcpbenchopen}, whose first version (v0.1.0) contains a collection of 164 discrete combinatorial problems drawn from well-established sources in the research community, as detailed in Table~\ref{tab:sources}.
To achieve high quality and diversity, we selected problems from: a) the popular “benchmark library for constraints” CSPLib~\citep{csplib}, b) the examples repository of the CPMpy modelling framework~\citep{guns2019increasing}, c) the extensive repository of Håkan Kjellerstrand~\citep{hakan_data}, d) a modelling course-based set of problems~\cite{michailidis2024constraint}, and e) a dataset of complex operations research problems (ComplexOR) \citep{xiao2024chainofexperts}. Notably, we also gathered and integrated multiple data instances for 23 CSPLib problems (in total 167 different instances) to accommodate more robust model evaluation.

\begin{table}[ht]
    \caption{Sources for DCP-Bench-Open v0.1.0 which contains 164 problems in total. Selection was based on CPMpy model availability to facilitate evaluation (the ground-truth model is necessary).}
    \centering
    \begin{tabular}{>{\raggedright\arraybackslash}p{3cm}rp{11cm}}
        \toprule
        \textbf{Source} & \textbf{\#} & \textbf{Selected Problems} \\
        \midrule
        CSPLib & 39 & IDs: 1--3, 5--16, 18--19, 21--24, 26, 28, 32--34, 39, 41, 44, 49--50, 53--54, 56--57, 67, 74, 76, 84. \\
        \midrule
        CPMpy Examples & 16 & \textit{bus\_scheduling, minesweeper, packing\_rectangles, jobshop, knapsack, mario, n\_puzzle, RCPS, room, send, set\_game, sudoku, tsp, agatha, wolf, zebra}. \\
        \midrule
        Håkan K. & 80 & From \textit{3\_coins} until \textit{knights\_tour\_circuit} (alphabetically), excluding test files. \\
        \midrule
        Course & 18 & All problems from the repository. \\
        \midrule
        ComplexOR & 11 & All discrete problems from the repository. \\ 
        \bottomrule
    \end{tabular}
    \label{tab:sources}
\end{table}

Table~\ref{tab:dataset_comparison} highlights the diversity and complexity of DCP-Bench-Open compared to existing benchmarks.
It spans optimization and satisfaction problems, with a wide range of decision variables, constraints, and 349 distinct constraint relations. In particular, the number of constraints per problem ranges from 1 to 2463, and the number of variables from 2 to 716, offering a significantly broader scope than NL4Opt and LGPs (Table \ref{tab:dataset_comparison}).

\subsection{Problem Structure}
Each dataset problem can be viewed as a self-contained Python file containing all the necessary components for evaluating the modelling capabilities of LLMs, with the following structure:

\begin{enumerate}
    \item \textbf{Metadata:} The source(s) of the problem along with any additional relevant information.
    \item \textbf{Problem Description:} The original problem statement as provided by the source, with an additional print instruction that specifies the required output structure.
    \item \textbf{Input Data (optional):}
    \begin{itemize}
        \item \textit{Default Instance (Prompt):} One valid data instance is provided in plain Python syntax (using only built-in types) to the LLM to showcase the input format.
        \item \textit{Test Instances (Hidden):} A set of diverse hidden data instances (e.g., varying sizes, graph structures, or numerical ranges) may exist for each problem to evaluate model correctness and robustness under different parameters.
    \end{itemize}
    \item \textbf{Model:} The decision variables, constraints, and objectives (if any) that make up a runnable CP model of the problem, using the CPMpy modelling framework.
    \item \textbf{Solution Printing:} Solution printing code, following the requirements in the problem description.
\end{enumerate}

\noindent Importantly, only the \textbf{Problem Description} and (optionally) the \textbf{Default Instance} are provided to the LLM, the remaining components are used exclusively for evaluation.

\subsection{Evaluation Metrics}
\label{subsec:evaluation}

As the focus lies on problems that may have more than one optimal solution, it is mandatory to adopt an appropriate metric to evaluate the generated constraint models.
Related works have used constraint-level, model-level, and solution-level accuracy~\cite{ramamonjison_augmenting_2022,michailidis2024constraint}.
The first two metrics require mapping the decision variables of the generated model with those from the ground-truth model.
However, when modelling complex problems, multiple possible viewpoints and formulations of the decision variables are possible, thus making it highly challenging to map variables between the generated and ground-truth models in a correct but generic way.
This leads us to opt for \textbf{solution-level accuracy}, which measures the correctness of the solution obtained by executing the generated code; correctness is evaluated with respect to the ground-truth model, considering both feasibility and optimality.
Additionally, this metric allows for evaluating LLMs on \textbf{any} modelling framework desired, as long as a final solution can be printed.

Formally, let $\mathcal{P}$ be the set of $N$ problems in the benchmark. For each problem $p \in \mathcal{P}$, we utilize a set of test data instances $\mathcal{D}_p = \{d_1, \dots, d_{k}\}$, where $d_1$ represents the default instance provided in the prompt.
Let $\hat{\mathbf{a}}_{p,d}$ be the solution produced by executing the LLM-generated model for problem $p$ on data instance $d$, and let $\mathbf{C}_{p,d}$ be the ground-truth constraint model instantiated with $d$.
We define an indicator function $\mathbb{I}_{correct}(p, d)$ which evaluates to 1 if the generated solution satisfies all constraints and (only for optimization problems) achieves the optimal objective value $f^*(\cdot)$:
\begin{equation}
\mathbb{I}_{correct}(p, d) = \mathbb{I} \left( \hat{\mathbf{a}}_{p,d} \in \text{Sol}(\mathbf{C}_{p,d}) \wedge f(\hat{\mathbf{a}}_{p,d}) = f^*(\mathbf{C}_{p,d}) \right)
\end{equation}

\noindent To account for the potential imbalance in the number of available data instances across problems, we define three solution-based metrics:

\paragraph{1. Single Instance Accuracy (SIA)} Measures whether the generated constraint model correctly solves the default data instance provided in the context. This serves as an optimistic upper bound on performance.
\begin{equation}
\text{SIA} = \frac{1}{N} \sum_{p \in \mathcal{P}} \mathbb{I}_{correct}(p, d_1)
\end{equation}

\paragraph{2. Multiple Instance Accuracy (MIA)} A strict metric requiring the generated model to correctly solve \textbf{all} test instances for a given problem. This penalizes generated models that overfit to the prompt's default example instance.
\begin{equation}
\text{MIA} = \frac{1}{N} \sum_{p \in \mathcal{P}} \left( \prod_{d \in \mathcal{D}_p} \mathbb{I}_{correct}(p, d) \right)
\end{equation}

\paragraph{3. Averaged Instance Accuracy (AIA)} Calculates the accuracy per problem first (for all problem's instances), then averages across problems. This ensures that problems with large instance sets available do not dominate the metric.

\begin{equation}
\text{AIA} = \frac{1}{N} \sum_{p \in \mathcal{P}} \left( \frac{1}{|\mathcal{D}_p|} \sum_{d \in \mathcal{D}_p} \mathbb{I}_{correct}(p, d) \right)
\end{equation}

\noindent In our experiments, we use SIA and evaluate the default instance, unless stated otherwise.

\section{Experiments}
\label{sec:experiments}

In our experiments, we focus on the following questions:

\begin{enumerate}

    \item[\textbf{Q1:}]
    How do LLMs perform in generating constraint models across different modelling frameworks?
    
    \item[\textbf{Q2:}]
    What types of errors appear in the generated constraint models, and how do different system prompts address them? How does self-verification affect runtime errors?
    
    \item[\textbf{Q3:}]
    What is the impact of inference-time computation methods, such as reasoning, retrieval-augmented in-context learning, repeated sampling, and self-verification, individually and in combination?
    
    \item[\textbf{Q4:}]
    How robust are the generated models when evaluating over multiple hidden data instances per problem?

    \item[\textbf{Q5:}]
    How do LLMs perform in generating constraint models across different modelling frameworks when using inference-time compute methods?

\end{enumerate}

\subsection*{Experimental Setup}
\label{subsec:experimental_setup}

Table~\ref{tab:llm_abbrev} lists the LLMs used throughout the experiments.
We considered covering a range of properties (coding, rank, mixture of experts, number of parameters, open weights/proprietary, context window size, etc.).
Specifically, we selected 7 models from the LMSys Chatbot Arena leaderboard based on their size\footnote{For instance, Mistral-Small-24B-Instruct-2501 reached only up to 20\% accuracy} and ranking on coding tasks~\cite{chiang2024chatbot}.\footnote{As accessed on 28 November 2025, \url{https://lmarena.ai/leaderboard}}
We used\footnote{The code is available here: \url{https://github.com/kostis-init/CP-Bench}} the
\textit{OpenAI}\footnote{\url{https://platform.openai.com/docs/api-reference}},
\textit{Together AI}\footnote{\url{https://docs.together.ai/reference}},
\textit{Anthropic}\footnote{\url{https://docs.anthropic.com/en/api}},
and \textit{DeepSeek}\footnote{\url{https://platform.deepseek.com/docs}}
APIs via their Python client libraries, using Python 3.12.
We set the reproducibility seed to 42 for all API calls, and the temperature parameter $\tau$ to 0 for deterministic outputs, except when applying repeated sampling techniques, where we increased it to $\tau = 0.8$ for variability, following existing works~\cite{olausson2023linc}. Notably, some LLMs do not support setting the temperature parameter (e.g. gpt-5.1), thus relying on their internal variability for sampling.
Additionally, we set output token limit per answer to 12k.

For executing the generated constraint models, we used a 10-second timeout due to the small instance sizes of the problems.
Specifically, all data instances selected for the multiple instance experiments (Q4 and Q5) are solved under 5 seconds using the ground-truth model. For the final experiment, we increased the generated model time limit to 30 minutes, to account for instances that were slower to solve with some frameworks. Finally, all experiments were conducted on an Ubuntu 24.04.3 system with 32GB RAM and an Intel Core Ultra 7 165Hx22 processor.

\begin{table}[ht]
    \caption{LLMs used in the experiments. Parameter counts and context sizes are approximate, active parameters for Mixture-of-Experts models are denoted in parentheses, and N/A means we found no public information about it.}
    \centering
    \begin{tabular}{l c c c c}
        \toprule
        \textbf{LLM} & \textbf{Organization} & \textbf{\# Parameters (Active)} & \textbf{Context Size} & \textbf{API Provider} \\
        \toprule
        gpt-5.1-2025-11-13       & OpenAI   & N/A     & 400k   & OpenAI         \\
        gpt-oss-120B       &   OpenAI      & 117B (5.1B)    & 200k & Together AI \\
        DeepSeek-V3.2         & DeepSeek & 685B (37B)    & 128k  & DeepSeek          \\
        Qwen3-Coder-480B-A35B        & Alibaba  & 480B (35B)     & 256k &    Together AI      \\
        Qwen3 235B A22B Instruct 2507   &      Alibaba    & 235B (22B)     & 262k & Together AI \\
        Kimi K2 Instruct     &  Moonshot AI   & 1T (32B)     & 256k & Together AI  \\
        Cogito v2.1 671B     &   Deep Cogito  & 671B (37B)    & 128k   & Together AI \\
        \bottomrule
    \end{tabular}
    \label{tab:llm_abbrev}
\end{table}



\subsection*{Q1: Modelling Frameworks}

\begin{figure*}[ht]
    \centering
    \includegraphics[width=0.9\linewidth]{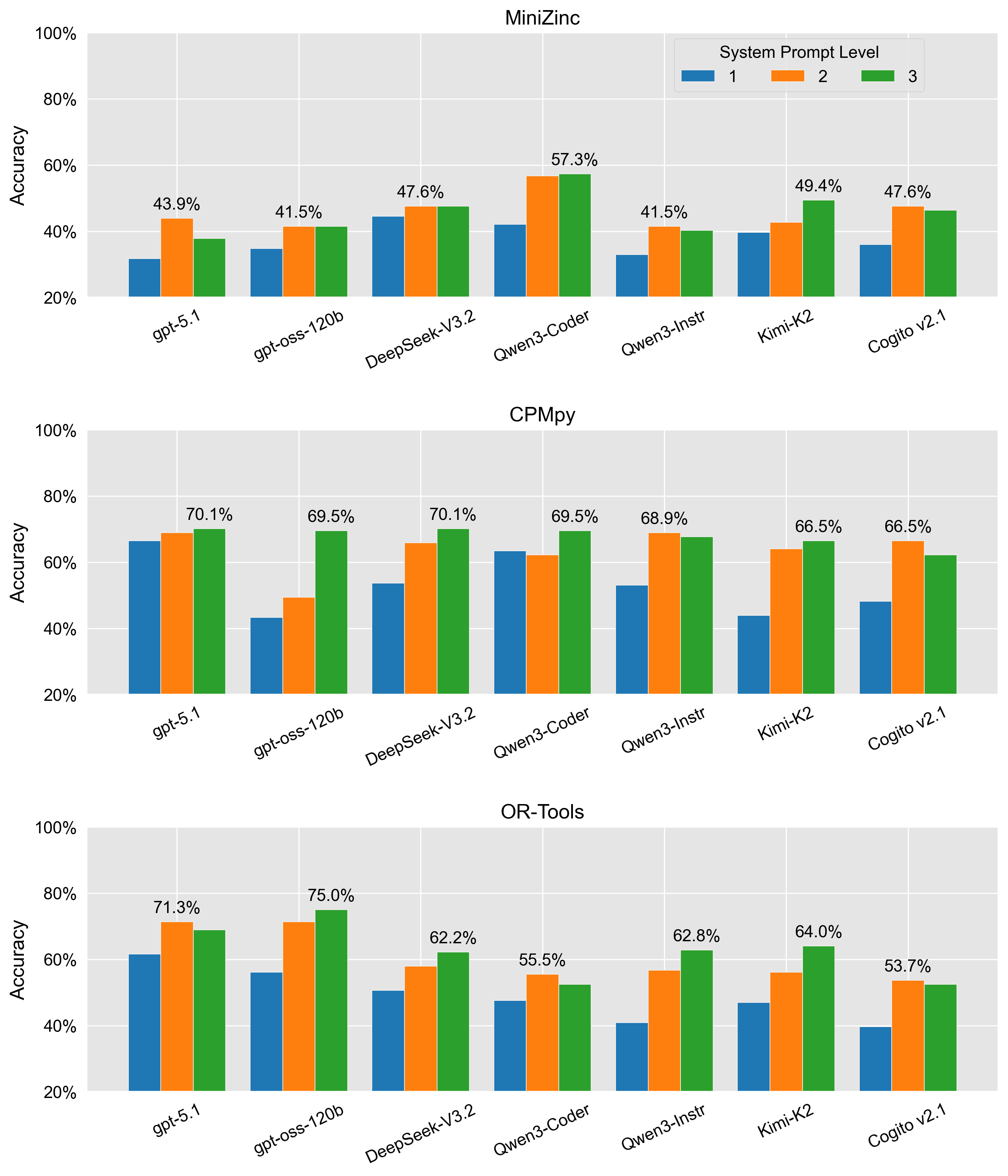}
    \Description{Three stacked bar charts comparing accuracy across MiniZinc (top), CPMpy (middle), and OR-Tools (bottom). The X-axis lists seven LLMs (including gpt-5.1, DeepSeek-V3.2, Qwen3 and others). The Y-axis represents Accuracy from 20\% to 100\%. Bars are grouped by System Prompt Level (1, 2, 3). General trends show that CPMpy and OR-Tools achieve higher accuracy (often 60-75\%) compared to MiniZinc (mostly 30-60\%), and System Prompt Levels 2 and 3 consistently outperform Level 1.}
    \caption{Percentages of successfully generated models (Single Instance Accuracy). From top to bottom: MiniZinc, CPMpy, OR-Tools.}
    \label{fig:all_frame_detail}
\end{figure*}

\begin{figure}[ht]
    \centering
    \includegraphics[width=0.7\linewidth]{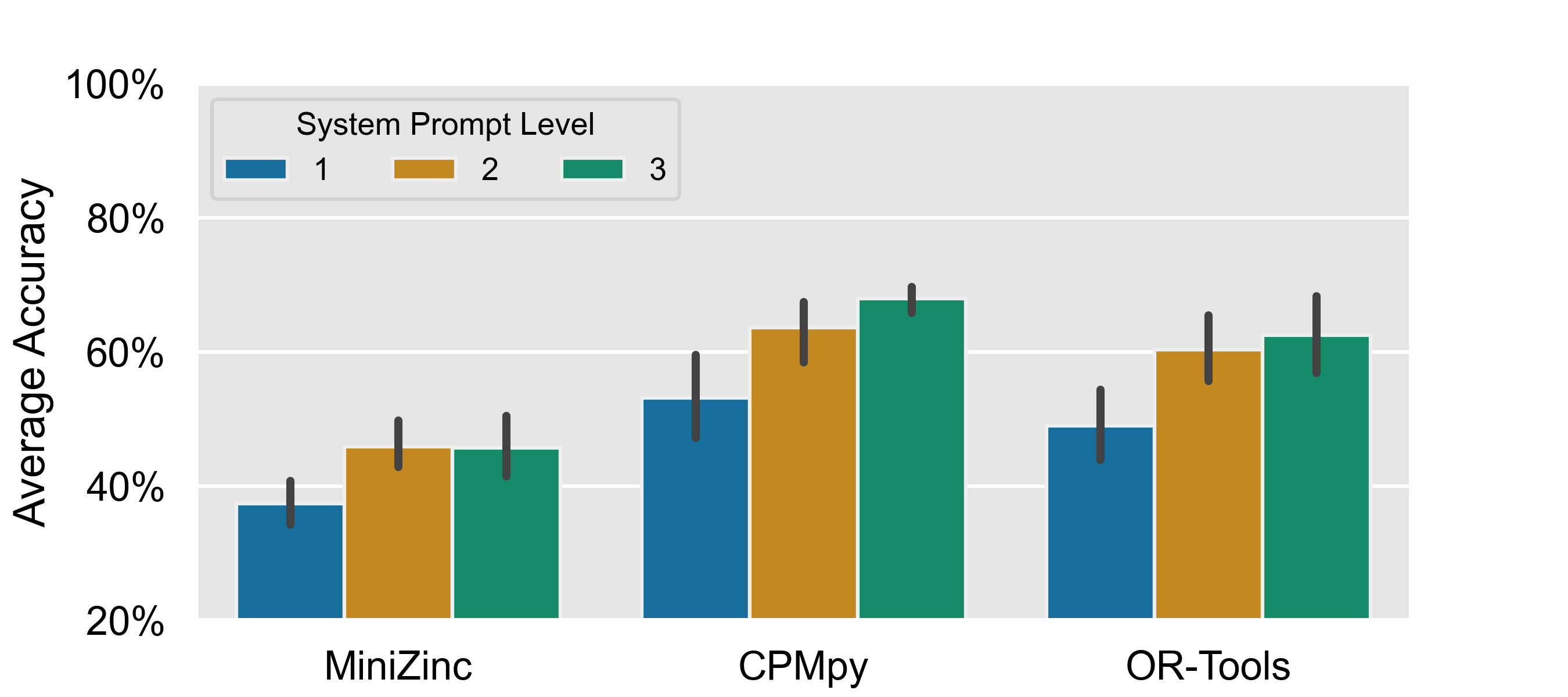}
    \Description{A bar chart showing the Average Accuracy (20\% to 100\%) for three frameworks: MiniZinc, CPMpy, and OR-Tools. Each framework has three bars corresponding to System Prompt Levels 1, 2, and 3. The trend shows that accuracy consistently improves as the prompt level increases. CPMpy achieves the highest overall accuracy (peaking near 70\% at Level 3), followed by OR-Tools, while MiniZinc remains below 50\% across all levels. Error bars indicate variability.}
    \caption{Average Single Instance Accuracy by Framework and System Prompt Level (Aggregated across LLMs).}
    \label{fig:all_frame_avg}
\end{figure}

Figure~\ref{fig:all_frame_detail} presents the detailed accuracy breakdown, while Figure~\ref{fig:all_frame_avg} illustrates the aggregated results across all LLMs for each modelling framework (MiniZinc, CPMpy, and OR-Tools CP-SAT) under different system prompt levels.

The aggregated results indicate that the choice of modelling framework significantly influences LLM success rates. When instructed to utilise CPMpy, LLMs exhibit consistently higher performance compared to MiniZinc and (slightly) OR-Tools, particularly when supplied with detailed system prompts (Levels 2 and 3). 
This gap is likely attributable to two factors.
First, contemporary LLMs are trained on vast corpora of Python code~\cite{chen2021evaluating}; this familiarity with Python syntax facilitates the generation of syntax error-free constraint models in Python-based frameworks (CPMpy and OR-Tools).
Second, CPMpy offers higher-level modelling abstractions—such as nested expressions, double reification, and a large set of built-in global constraints and global functions, which simplify the translation from natural language specifications to code.

Analyzing the individual LLM performance across frameworks (Figure~\ref{fig:all_frame_detail}) reveals distinct patterns for each framework:

\textbf{MiniZinc:} The code-specialized \texttt{Qwen3-Coder} significantly outperforms other models, achieving a peak accuracy of 57.3\%. Notably, \texttt{Qwen3-Coder} and \texttt{Kimi-K2-I} are the only models that gain substantial benefit from the inclusion of documentation (Level 3). For other LLMs, added documentation fails to increase accuracy. This suggests that for domain-specific languages like MiniZinc, which also have limited presence in pre-training data, code-specialized LLMs with explicit syntax documentation yield higher accuracies.

\textbf{CPMpy:} In contrast, results for CPMpy show higher consistency. Performance is evenly distributed, with a narrow variance between the lowest performing models (\texttt{Kimi} and \texttt{Cogito} at 66.5\%) and the top performers (\texttt{gpt-5.1} and \texttt{DeepSeek-V3.2} at roughly 70.1\%). This consistency is present only under the documentation prompt (Level 3), at which LLM choice seems to become a less critical factor than in other frameworks.

\textbf{OR-Tools:} OpenAI's LLMs are leading with \texttt{gpt-oss-120b} achieving the highest overall accuracy (75.0\%), closely surpassing \texttt{gpt-5.1} (71.3\%). Conversely, \texttt{Qwen3-Coder}, despite its dominance in MiniZinc, struggles significantly with OR-Tools (max $\sim$55.5\%). Interestingly, this also comes in contrast with \texttt{gpt-oss-120b}, which excels at OR-Tools but vastly underperforms in MiniZinc.


In summary: a) generating constraint models using Python libraries tends to yield higher accuracy than the domain-specific MiniZinc language, and b) the documented high-level framework (CPMpy with documentation prompt) show the highest consistency in accuracy over all LLMs tested (>60\%).

\subsection*{Q2: System Prompt \& Error Types}

Figures~\ref{fig:all_frame_detail} and \ref{fig:all_frame_avg} also demonstrate the impact of varying system prompt levels across frameworks and LLMs.
A key observation across all configurations is that providing at least some specific guidance — either classroom-level guidelines (Level 2) or framework documentation (Level 3) — consistently yields higher accuracy than the basic prompt (Level 1).
However, the addition of detailed documentation on top of the guidelines showed some variations.
For the Python-based frameworks (CPMpy, OR-Tools), API documentation in the system prompts is beneficial to produce valid constraint models, as it led to an increase in accuracy for 5 and 4 (respectively) out of the 7 LLMs tested.
Similarly, 
Interestingly, for MiniZinc, while adding guidelines (Level 2) improved performance over the baseline (Level 1) for all LLMs, further adding detailed documentation (Level 3) did not consistently improve upon Level 2.
In fact, average performance slightly decreased, with only 2 out of 7 LLMs showing any benefit from the extra documentation on MiniZinc.

\begin{figure}[ht]
    \centering
    \includegraphics[width=0.99\linewidth]{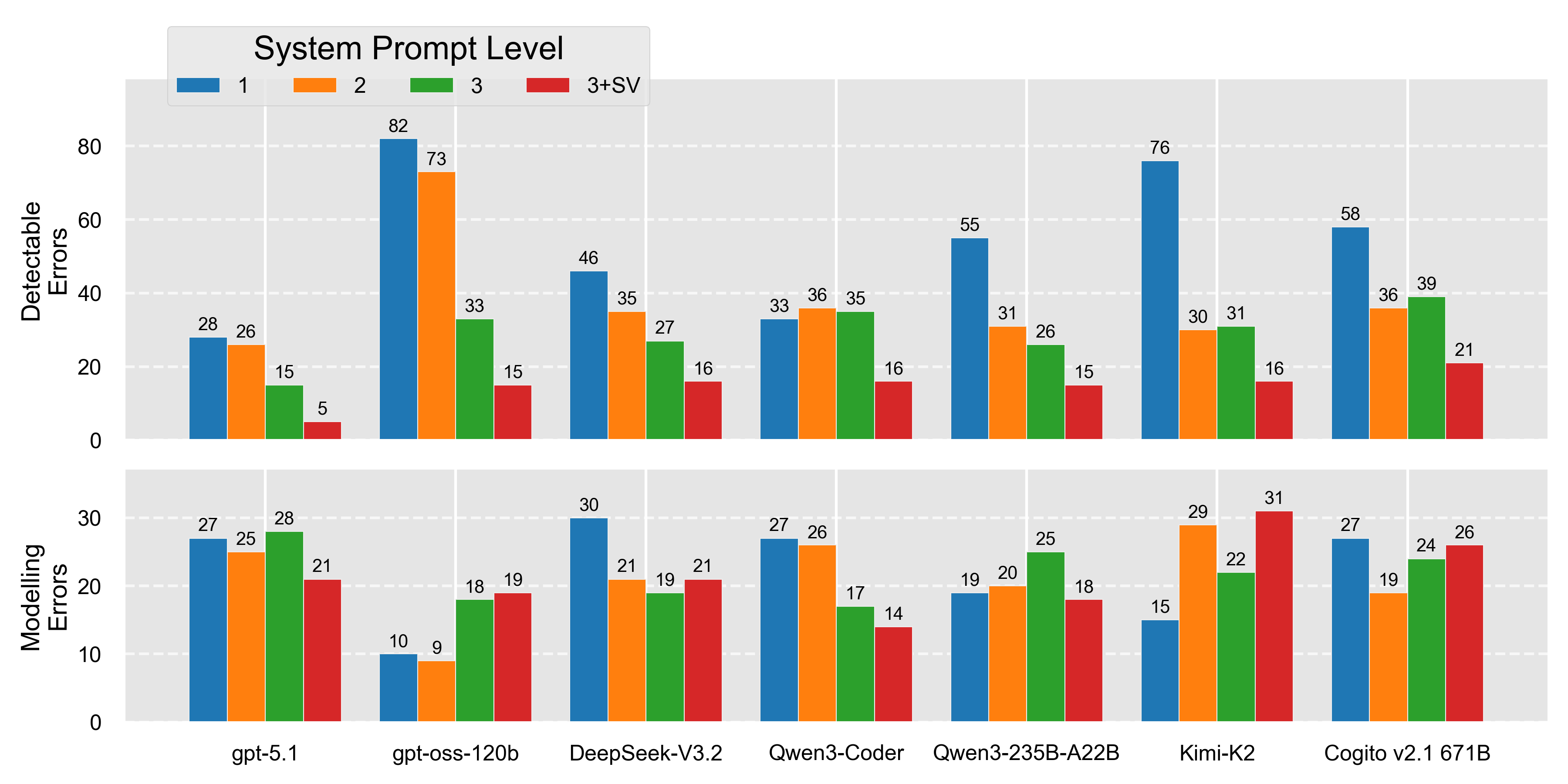}
    \Description{Two vertically stacked bar charts comparing error counts for different LLMs (x-axis) across System Prompt Levels 1, 2, 3, and 3+SV (colored bars). The top chart ("Detectable Errors") shows a strong downward trend: as the prompt level increases and self-verification is added (red bars), detectable errors drop significantly (e.g., GPT-5.1 drops from 28 to 5). The bottom chart ("Modelling Errors") shows mixed results, with counts remaining relatively stable or fluctuating slightly across prompt levels.}
\caption{Detectable and modelling errors with CPMpy. 3+SV stands for Self-Verification of a single repetition along with a system prompt level 3 (documentation). Importantly, the (green) bars for level 3 represent the same experimental runs as 3+SV (red), but report the errors prior to the self-verification iteration, for reproducibility.}
    \label{fig:detectable_cpmpy}
\end{figure}

To understand \textit{why} models fail, we categorize failures into two distinct types: \textbf{Detectable Errors} (syntax errors, invalid API usage, or running into timeouts) and \textbf{Modelling Errors} (the code executes but yields incorrect solutions or unsatisfiable models).
For CPMpy, depicted in Figure~\ref{fig:detectable_cpmpy}, as the prompt level increases, detectable errors decrease while modelling errors rise.
A similar trend was observed for MiniZinc and OR-Tools (figures are available in Appendix~\ref{sec:app_errors}).
The reduction in detectable errors with more detailed prompts suggests that LLMs generate more syntactically correct and executable code as they receive more information. This partly leads to more successes, but at times also to an increase in logical modelling errors (bottom subfigure), where the code is syntactically correct but the solution is either incorrect or not found.

\paragraph{Self-Verification}

To further reduce the number of detectable errors, we also report a comparison with Self-Verification (SV) being used on top of the documentation prompt (Figure~\ref{fig:detectable_cpmpy}, label 3+SV). Results show the positive effect of Self-Verification, even on only 1 single repetition. Across nearly all LLMs, SV significantly reduces detectable errors compared to a pure documentation-rich prompt. For instance, \texttt{gpt-5.1} reduces detectable errors from 19 to 5, and \texttt{Qwen3-Coder} from 30 to 16. This indicates that LLMs are capable of interpreting error tracebacks to fix their own syntax and API misuse.
However, the impact on modelling errors is mixed. For stronger models like \texttt{gpt-5.1} and \texttt{Qwen3-Coder}, SV reduces both error types, suggesting full improvement, but for models like \texttt{Kimi-K2-I}, SV reduces detectable errors (from 37 to 16) but nearly doubles modelling errors (from 17 to 31). These results indicate a correlation between the reduction in detectable errors and the increase in modelling errors, where the initial runtime error may be corrected but nonetheless result in a logical modelling error.

\subsection*{Q3: Inference-Time Compute Methods}

We assess the impact of various inference-time compute strategies introduced in Section~\ref{sec:methodology}, with the following configurations:

\begin{enumerate}
\item{\textbf{Baseline} (Section~\ref{subsec:system_prompt}):} The input consists of the system prompt and problem description. The first generated model is directly evaluated without further processing, as in Q1.
\item{\textbf{RAICL} (Section~\ref{subsec:raicl}):} The input contains the system prompt and $e=8$ examples based on reverse semantic similarity, following existing work \cite{michailidis2024constraint}. We use a leave-one-out strategy for evaluation, testing each problem individually while utilizing the 100 remaining ones as the retrieval database.
\item{\textbf{Reasoning}:} We investigate internal "reasoning" as a test-time method for improving model generation. Depending on the model, this involves either prompting for "Low Reasoning Effort" (gpt-5.1, gpt-oss, Qwen3-Coder) or utilizing a reasoning-specialized variant (e.g., DeepSeek-Reasoner, Kimi-Thinking).
\item{\textbf{Sampling} (Section~\ref{subsec:repeated_sampling}):} Generation of $k=10$ responses in parallel, with the final model selected via solution majority voting. The input consists of the system prompt and problem description.
\item{\textbf{Self-Verification} (Section~\ref{subsec:self-debug}):} Iterative refinement of the Baseline generation through the self-verification loop, with a maximum of 10 iterations.
\item{\textbf{Sampling \& Self-Verification}:} A hybrid approach where the output selected by Sampling is subsequently refined via Self-Verification. Again, we use 10 samples and maximum 10 iterations.
\end{enumerate}

\begin{figure}[ht]
    \centering
    \includegraphics[width=0.99\linewidth]{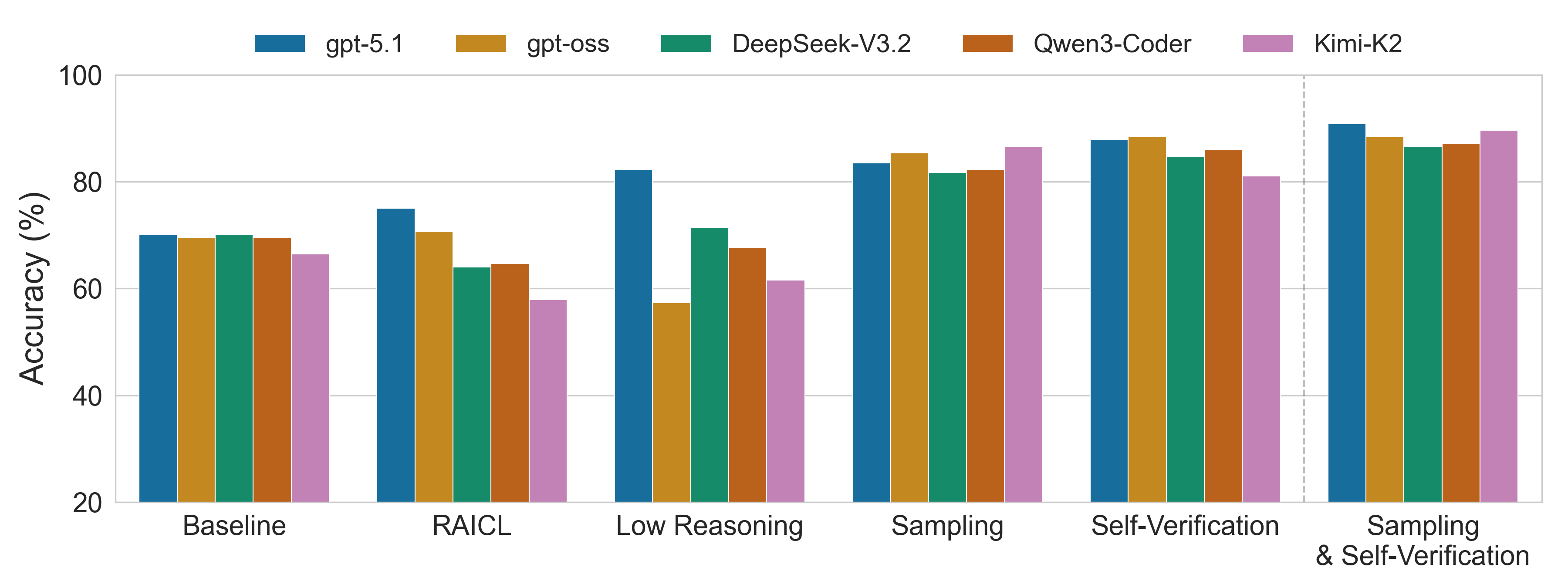}
    \Description{A bar chart comparing the Accuracy (\%) of five LLMs (GPT-5.1, GPT-OSS, DeepSeek-V3.2, Qwen3-Coder, Kimi-K2) across inference strategies: RAICL, Low Reasoning, Sampling, Self-Verification, and Sampling & Self-Verification. The trend shows that advanced inference-time methods significantly boost performance. While the Baseline sits around 65-70\%, "Sampling" and "Self-Verification" push accuracy above 80\%. The highest performance is achieved by "Sampling & Self-Verification" (hatched bars), reaching approximately 90\% across most LLMs.}
    \caption{Performance (Single Instance Accuracy) across different inference-time compute methods for each LLM using CPMpy.}
    \label{fig:methods_succ_err}
\end{figure}

\noindent We included five LLMs (those that benefit from the Level 3 Documentation prompt): \textit{gpt-5.1}, \textit{gpt-oss}, \textit{DeepSeek-V3.2}, \textit{Qwen3-Coder}, \textit{Kimi-K2}, and we used CPMpy due to its higher accuracy on average (Q1). Based on the Q2 results, we used the documentation system prompt (Level 3). The results are summarized in Figure~\ref{fig:methods_succ_err}.

\paragraph{Baseline \& RAICL} Surprisingly, the RAICL strategy was shown ineffective for most LLMs or occasionally slightly effective. For most models, including DeepSeek-V3.2, Qwen3-Coder, and Kimi-K2, performance degraded when retrieval-augmented in-context examples were provided compared to the zero-shot Baseline.
Based on these results and compared to the results from~\citet{michailidis2024constraint}, it seems that for formal constraint modelling, LLMs may utilize the structured framework documentation in the system prompt more effectively than they generalize from retrieved, problem-specific specific model examples provided in context.
This finding also can simplify practical deployment by removing the need for custom example databases and dynamic retrieval methods.
As such, investigating more modelling tools or frameworks could start on merely adapting their documentation for the system prompt.

\paragraph{Reasoning} The "Low Reasoning" strategy showed high variance accross LLMs. While \textit{gpt-5.1} showed substantial gains, suggesting its internal reasoning chain benefits constraint formulation, other models such as \textit{gpt-oss} and \textit{Kimi-K2} saw roughly 5-10\% performance drops. As such, "reasoning" LLMs on their own are not yet an absolute enhancement for formal modelling.

\paragraph{Sampling \& Self-Verification} These methods boosted performance across all LLMs (more than 10\% increase in accuracy), with the most significant gains observed by Kimi-K2, which increased accuracy by roughly 20\% compared to the Baseline.
Notably, with repeated sampling, open-source LLMs managed to slightly outperform OpenAI's larger LLM (gpt-5.1).
Self-verification also improved accuracy significantly compared to the Baseline, yielding results comparable to repeated sampling, although Kimi-K2 performed slightly better with sampling than with self-verification alone.
As RAICL and Reasoning failed to improve accuracy consistently across all LLMs, we experimented with the combination of the two best-performing methods: repeated sampling \& self-verification.
This combined approach further increased accuracy, with all LLMs tested demonstrating their highest performance in this setting, reaching up to 91\% accuracy (gpt-5.1); only 2 instances resulted in detectable errors (1 runtime error, 1 wrong solution format generation), and 13 instances resulted in modelling errors (unsatisfiable or wrong solution).


\subsection*{Q4: Multiple Data Instances}

\begin{figure}[ht]
    \centering
    \includegraphics[width=0.9\linewidth]{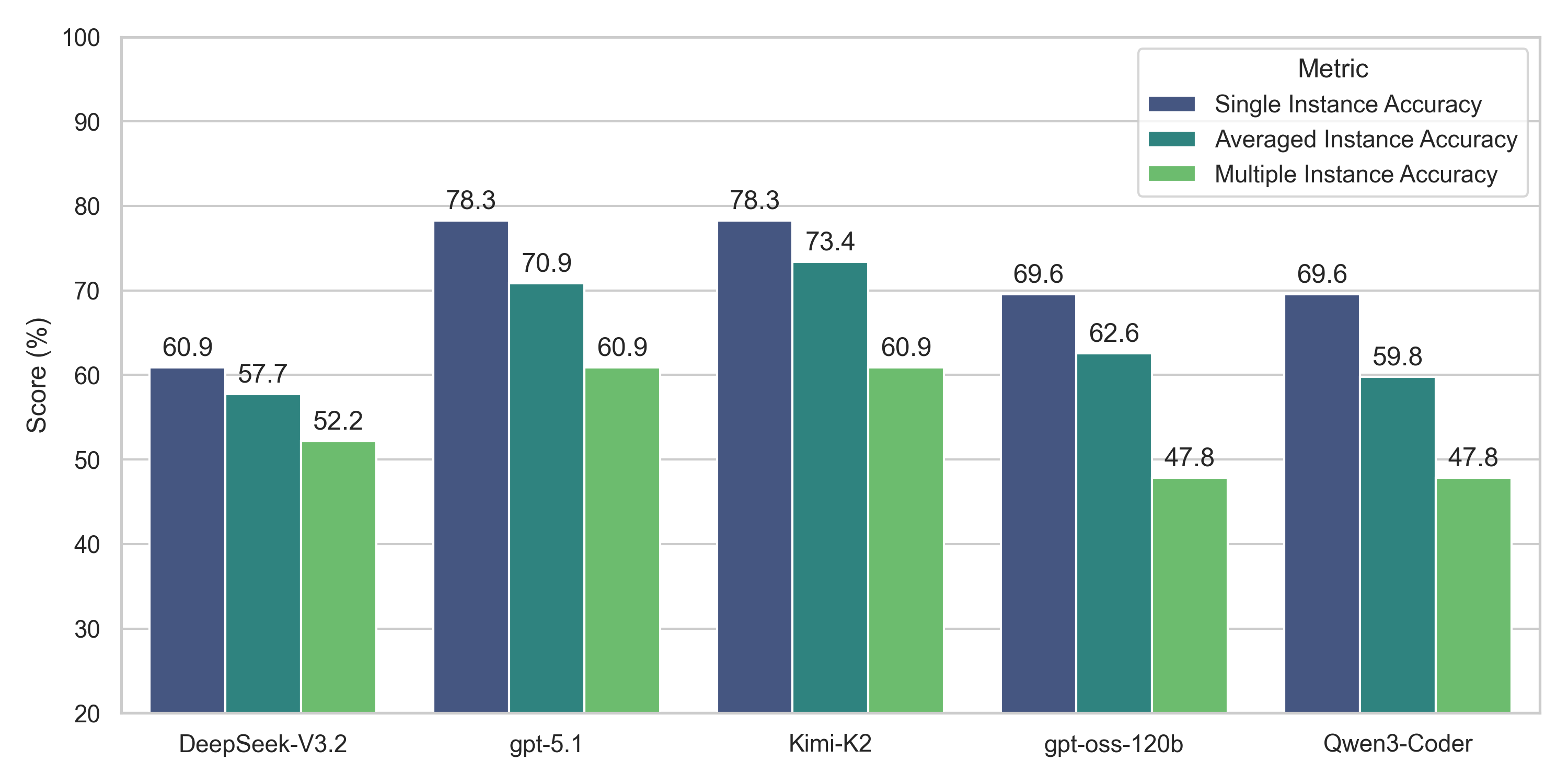}
    \Description{A grouped bar chart comparing three reliability metrics across five LLMs (DeepSeek-V3.2, gpt-5.1, gpt-oss-120b, Qwen3-Coder, Kimi-K2-I). The three metrics are Single Instance Accuracy (dark blue), Averaged Instance Accuracy (teal), and Multiple Instance Accuracy (green). The chart shows a consistent "stair-step" drop in performance for every model: Single Instance Accuracy is always highest (e.g., gpt-5.1 at 78.3\%), while Multiple Instance Accuracy is always lowest (e.g., gpt-5.1 drops to 60.9\%). This highlights that solving all hidden instances (strict consistency) is harder than solving just the default prompt example.}
    \caption{Comparison of model reliability metrics when all instances per problem are evaluated (only using problems that have more than 1 data instance). \textbf{Single Instance Accuracy} (optimistic) measures success on the prompt's default example. \textbf{Multiple Instance Accuracy} (strict) requires success on all hidden instances.}
    \label{fig:reliability_metrics}
\end{figure}

To evaluate model generalization and robustness, we compared the performance of generated models on the default instance provided in the prompt versus a set of hidden test instances.
In this experiment, the input prompt explicitly defines the expected data structure and parameter types by including the default instance. Crucially, the prompt instructs the LLM to generate a model that assumes these parameter values are pre-loaded into the namespace, thereby allowing to use the same generated model for all the hidden instances of the problem.
We restricted this evaluation to only problems containing multiple data instances, resulting in a subset of 23 problems with 167 total instances. Figure~\ref{fig:reliability_metrics} shows the gap between Single Instance Accuracy (SIA), Averaged Instance Accuracy (AIA), and Multiple Instance Accuracy (MIA).

Across all evaluated LLMs, we observe a consistent performance decline when moving from the single default instance to the full instance set. While gpt-5.1 and Kimi-K2-I achieve the highest performance across all metrics (SIA: 78.3\%, MIA: 60.9\%), they both still show a 17.4\% drop in strict robustness. Notably, despite gpt-oss-120b and Qwen3-Coder achieving an SIA of 69.6\%, their MIA drops to 47.8\%, a relative decrease of over 30\%. The smallest decline was observed with DeepSeek-V3.2, which dropped by only 8.7\%.
These results suggest that LLMs frequently overfit to the specific values or structure of the default data instance provided in the prompt (e.g., hardcoding array sizes or assuming specific domain ranges) rather than modelling the abstract constraints of the problem. Multiple Instance Accuracy can indeed serve as a stricter correctness metric in this domain.

\subsection*{Q5: Multi-Instance Evaluation with Inference-Time Compute}

Finally, we integrate our findings by evaluating the strongest configuration—Sampling ($k=10$) combined with Self-Verification ($i=10$) using the Documentation Prompt—on the full dataset across all modelling frameworks. As in the previous experiment (Q4), problem data is loaded externally rather than being hardcoded in the prompt.

\begin{table}
    \caption{Impact of inference-time compute on Single-Instance Accuracy across frameworks. The Baseline uses the Level 3 documentation prompt with problem data included directly in the prompt context. The Inference-Time Compute (ITC) method employs Repeated Sampling ($k=10$) combined with Self-Verification (max 10 iterations)-as in the combined method in Q3-and loads problem data externally.}
    \centering
    \begin{tabular}{l c c c c}
        \toprule
        & \multicolumn{2}{c}{\textbf{gpt-5.1-2025-11-13}} & \multicolumn{2}{c}{\textbf{Kimi K2 Instruct}} \\
        \cmidrule(lr){2-3} \cmidrule(lr){4-5}
        \textbf{Framework} & Baseline & Sampling \& Self-Verification & Baseline & Sampling \& Self-Verification \\
        \midrule
        MiniZinc & 37.8\% & \textbf{73.2\%} & 49.4\% & \textbf{72.0\%} \\
        CPMpy    & 70.1\% & \textbf{90.2\%} & 66.5\% & \textbf{89.0\%} \\
        OR-Tools & 68.9\% & \textbf{88.4\%} & 64.0\% & \textbf{82.3\%} \\
        \bottomrule
    \end{tabular}
    \label{tab:ttc_with_all_frameworks_comparison}
\end{table}

\begin{figure}
    \centering
    \includegraphics[width=0.9\linewidth]{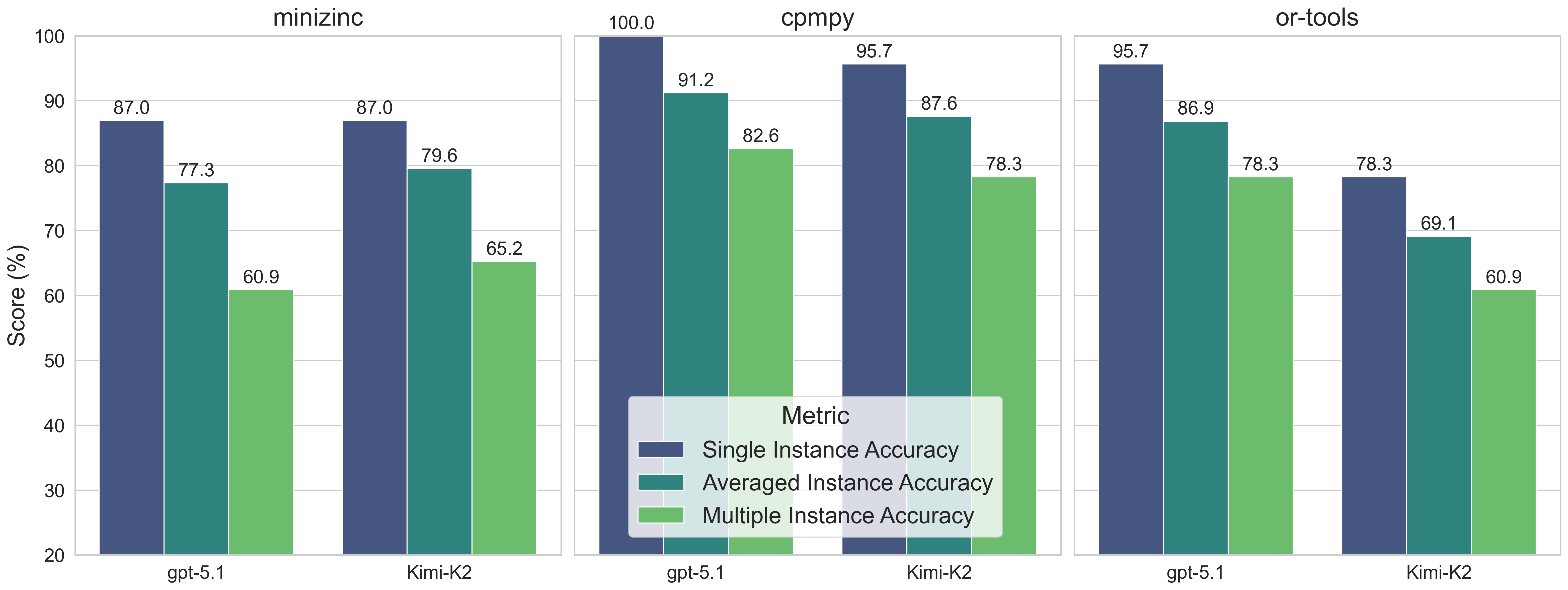}
    \Description{Three side-by-side bar charts displaying accuracy percentages for MiniZinc, CPMpy, and OR-Tools. Each chart compares two LLMs (gpt-5.1 and Kimi-K2-I) across three metrics: Single Instance Accuracy (blue), Averaged Instance Accuracy (teal), and Multiple Instance Accuracy (green). A clear decline is visible in all charts: performance is highest on Single Instance Accuracy and lowest on Multiple Instance Accuracy. CPMpy generally yields the highest scores, with gpt-5.1 achieving 100\% Single Instance and over 80\% Multiple Instance Accuracy. OR-Tools follows closely, while MiniZinc shows slightly lower performance, dropping to around 60-65\% for the strict Multiple Instance metric.}
    \caption{Comparison of accuracy for all frameworks and all multi-instance problems (23) when all instances per problem are evaluated using the combined Sampling \& Self-Verification method with the Documentation Prompt. For this experiment, a timeout of 30 minutes was set for executing each generated model. From left to right: MiniZinc, CPMpy, OR-Tools.}
    \label{fig:combined_frameworks}
\end{figure}

As shown in Table~\ref{tab:ttc_with_all_frameworks_comparison}, \textbf{applying inference-time compute yields impressive improvements over the baseline across all frameworks}. For CPMpy, Single Instance Accuracy increases from 70.1\% to 90.2\% for \texttt{gpt-5.1}, and from 66.5\% to 89.0\% for \texttt{Kimi-K2}. Similar gains are observed for OR-Tools ($\sim$20\% increase) and MiniZinc ($\sim$35\% increase for gpt-5.1).

Regarding multi-instance robustness, Figure~\ref{fig:combined_frameworks} shows a similar trend to Q4, where there are still several occurrences of models that correctly solve the default instance but fail on all hidden ones. As in Q4, in this experiment we again only included the 23 problems that contain more than one instance in the dataset, and we set the timeout limit to 30 minutes. Notably, both the highest MIA and the smallest relative decline is observed when modelling with CPMpy, closely followed by OR-Tools with gpt-5.1, and MiniZinc with Kimi-K2. After close inspection upon the specific error types in hidden instances when the default one is solved correctly, we found that the majority of them for CPMpy and OR-Tools were that either the generated solution was incorrect or the model unsatisfiable. In contrast, for MiniZinc, code execution errors and wrong solutions were more common, followed by model timeouts (the detailed numbers are shown in Appendix~\ref{app:error}).

\section{Related Work}

\paragraph{Inference-Time Compute and Scaling}

Recent LLM literature has shifted from scaling model parameters during pre-training to scaling computational resources at inference time. \citet{snell2024scaling} demonstrated that optimally scaling test-time compute can allow smaller models to outperform significantly larger ones. \citet{muennighoff2025s1} categorizes test-time scaling into two dimensions: parallel and sequential scaling. Parallel scaling involves generating independent candidate solutions and selecting the most likely correct answer through methods like Majority Voting (or Self-Consistency decoding) \citep{wang2022self}, including also massive parallel sampling for solving competitive programming tasks (AlphaCode) \cite{li2022competition}. Sequential scaling, by contrast, allocates compute to iterative refinement, enabling LLMs to self-correct syntax and logical errors when provided with execution feedback from an interpreter or unit test results \citep{madaan2023self, chen2023teaching}. More recently, hybrid frameworks have emerged to combine these two dimensions, such as S*, which incorporates a self-debugging chain in each unique sample \citep{li-etal-2025-test}. Furthermore, reasoning-native models such as DeepSeek-R1 have internalized these processes through large-scale reinforcement learning, enabling backtracking and "thinking" tokens before producing a final response \citep{guo2025deepseek}.

\paragraph{LLMs for Formal Modelling and Optimization}

Early works on transforming natural language inputs into formal optimization models were motivated by the NL4Opt competition  \citep{ramamonjison_nl4opt_2023}, with various succesfull approaches on both entity recognition \citep{ner4opt_2023} and model generation \citep{prasath_synthesis_2023}.
More recent neuro-symbolic methods utilize LLMs to translate textual problem descriptions into various declarative representations for symbolic solvers, spanning Satisfiability (SAT) formulas \citep{ye_satlm_2023}, First-Order Logic \citep{olausson2023linc}, Answer Set Programming \citep{ishay_leveraging_2023}, Mixed-Integer Programming (MIP) \citep{ahmaditeshnizi2023optimus}, and PDDL for planning \citep{liu2023llm+}. Specifically for CP modelling, we have explored decomposition-based pipelines \citep{tsouros_holy_2023} and retrieval-augmented in-context learning with intermediate blueprint model representations \citep{michailidis2024constraint}. Additionally, \citet{shi-etal-2025-constraintllm} present a modelling approach based on supervised fine-tuned LLMs, while \citet{szeider2025cp} and \citet{cai2025gala} investigate agentic-based frameworks for CP modelling. Finally, distinct from pure modelling, \citet{voboril2025generating} introduce an LLM-based method to generate streamlining constraints that accelerate the solving process.

\paragraph{Existing Formal Modelling Benchmarks}

The evaluation of LLMs for optimization has primarily focused on Linear Programming (LP) and Mixed-Integer Programming (MIP). Early works like NL4Opt \citep{ramamonjison_nl4opt_2023} targeted entity extraction and formulation for small and homogeneous linear problems. More recent benchmarks such as NLP4LP \citep{ahmaditeshnizi2023optimus}, MAMO \citep{huang2024mamo}, IndustryOR \citep{tang2024orlm}, BWOR \citep{zhang2025or}, ComplexOR \citep{xiao2024chainofexperts}, and CO-Bench \citep{Sun2025COBench} expanded this scope to textbook and industrial operations research problems. To the best of our knowledge, in the domain of Constraint Programming, benchmarks remain limited. Parallel efforts include: CPEval \citep{song2025llmcp} and IndusCP \citep{shi-etal-2025-constraintllm} which offer datasets of classic problems derived from various well-established sources, and Text2Zinc \citep{singirikonda2025text2zinc} which introduces a cross-domain dataset containing CP, LP, and MIP problems with MiniZinc ground-truth models. Our DCP-Bench-Open complements these efforts by focusing specifically on discrete combinatorial problems, allowing for \textit{modelling framework-independent} multi-instance and multi-solution evaluation.

\section{Conclusion}
\label{sec:conclusion}

In this work, we collected a diverse dataset of discrete combinatorial problems and systematically evaluated various state-of-the-art LLMs on transforming textual problem descriptions into executable constraint models. We explored the use of different modelling frameworks, with varying abstraction and interface types.
We also explored the use of prompt-based and inference-time compute methods for enhancing the performance of LLMs in constraint modelling.

Our evaluation indicated an advantage of Python-based frameworks (CPMpy, OR-Tools) compared to the domain-specific MiniZinc language, with further improvements when appending detailed guidelines and documentation in system prompts.
Inference-time methods were shown to provide a significant improvement in general, with repeated sampling and self-verification showing up to a 35\% increase in accuracy (MiniZinc).
Notably, retrieval-augmented in-context learning was not consistently effective when combined with increased documentation in the prompts, and considering also its requirement for constructing example databases and selection strategies, suggests practical challenges for it in this domain.
The strongest performance came from combining repeated sampling and self-verification with the documentation prompt, achieving up to 90\% accuracy on CPMpy modelling, suggesting that LLMs can indeed assist in the modelling process. Our multi-instance evaluation does show that LLMs can over-encode the example instance given in the prompt, leading to incorrect results on other instances of the same problem type. Further work on multi-instance aware prompting, for example, multi-instance inference time computation could remedy this.

The existence of DCP-Bench-Open allows for systematically evaluating additional design choices, including newer LLMs, additional modeling or solving frameworks (e.g., different CP solvers, or directly modeling for integer linear programming solvers or SAT solvers), more coding LLM techniques (e.g. retrieval-augmented documentation), or supervised fine-tuning of LLMs. The latter would require a training dataset of constraint models and problem descriptions, in addition to the use of DCP-Bench-Open as evaluation dataset.
While our benchmark contains diverse and realistic combinatorial problems, they often originate from textbooks. Large-scale industrial problems typically involve more data, many constraints and objectives, and elaborate descriptions, though these are rarely publicly available.
Future work could also explore multi-turn interactions for modelling with LLMs, where a user and system iteratively refine the model through conversation and trial-and-error.
Finally, while this paper focused on evaluating model correctness, adding more instances in DCP-Bench-Open is straightforward. This opens the door to also evaluating modeling efficiency across multiple instances, allowing in turn to connect to LLM-driven multiple model generation, model selection based on efficiency as well as solver selection and algorithm configuration.

\begin{acks}
This project has received funding from the European Research Council (ERC) under the European Union’s Horizon 2020 research and innovation program (Grant No. 101002802, CHAT-Opt), and from the Flemish Government under “Onderzoeksprogramma Artificiële Intelligentie (AI) Vlaanderen”.
\end{acks}

\printbibliography

\appendix

\section{Error Analysis for Q5}
\label{app:error}

Table~\ref{tab:error_breakdown_wide} shows the absolute error numbers by LLM and Framework in the multi-instance inference-time compute experiment. Specifically, in this table we count the errors only when the default instance is solved correctly.

\begin{table}[h]
    \centering
    \caption{Detailed count of errors by LLM and Framework for the multi-instance ITC experiment (Q5) when the default instance is solved correctly.}
    \label{tab:error_breakdown_wide}
    \resizebox{\textwidth}{!}{%
    \begin{tabular}{l|ccc|ccc}
        \toprule
        & \multicolumn{3}{c|}{\textbf{gpt-5.1}} & \multicolumn{3}{c}{\textbf{Kimi-K2-I}} \\
        \textbf{Error Type} & \textbf{CPMpy} & \textbf{MiniZinc} & \textbf{OR-Tools} & \textbf{CPMpy} & \textbf{MiniZinc} & \textbf{OR-Tools} \\
        \midrule
        Wrong Solution & 4 & 7 & 4 & 5 & 3 & 15 \\
        Unsatisfiable & 6 & - & 6 & 6 & - & 1 \\
        Runtime Error & 1 & 6 & 1 & - & 5 & - \\
        Gen. Model Timeout & - & - & - & - & 5 & 4 \\
        \bottomrule
    \end{tabular}%
    }
\end{table}


    

\section{Detailed Errors for MiniZinc \& OR-Tools}
\label{sec:app_errors}

\begin{figure}[ht!]
    \centering
    \includegraphics[width=0.99\linewidth]{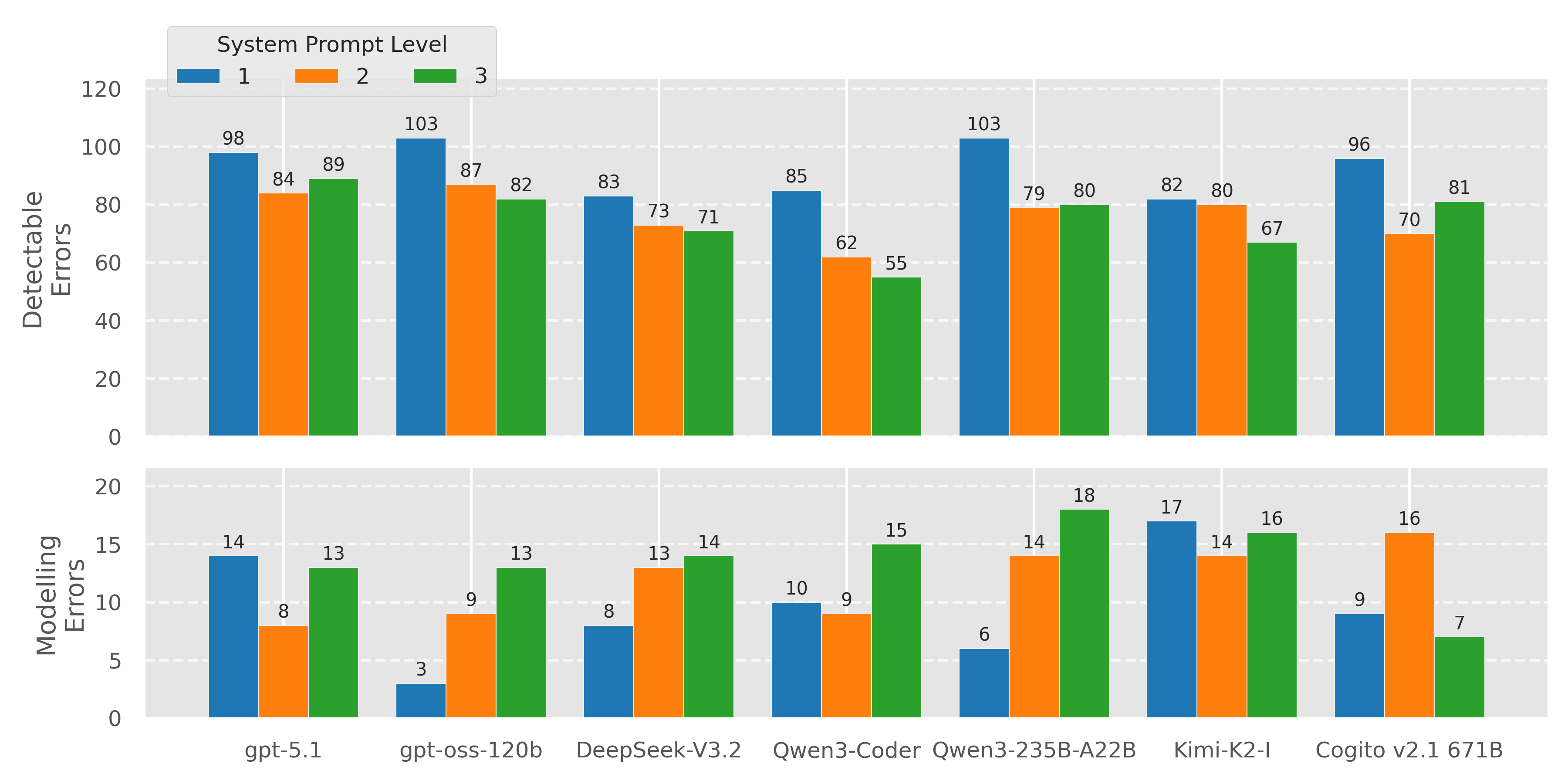}
    \Description{Two vertically stacked bar charts showing error counts for MiniZinc generation across seven LLMs. The top chart ("Detectable Errors") displays high error frequencies, with counts often ranging between 80 and 100 regardless of the System Prompt Level. The bottom chart ("Modelling Errors") uses a much smaller scale (0-20), showing that logical errors are rare (mostly under 15) compared to syntax errors. Increasing the prompt level (from 1 to 3) yields a consistent moderate reduction in detectable errors for MiniZinc.}
    \caption{Detectable and modelling errors when LLMs produce MiniZinc code through all system prompt levels.}
    \label{fig:mnz_err}
\end{figure}

\begin{figure}[ht!]
    \centering
    \includegraphics[width=0.99\linewidth]{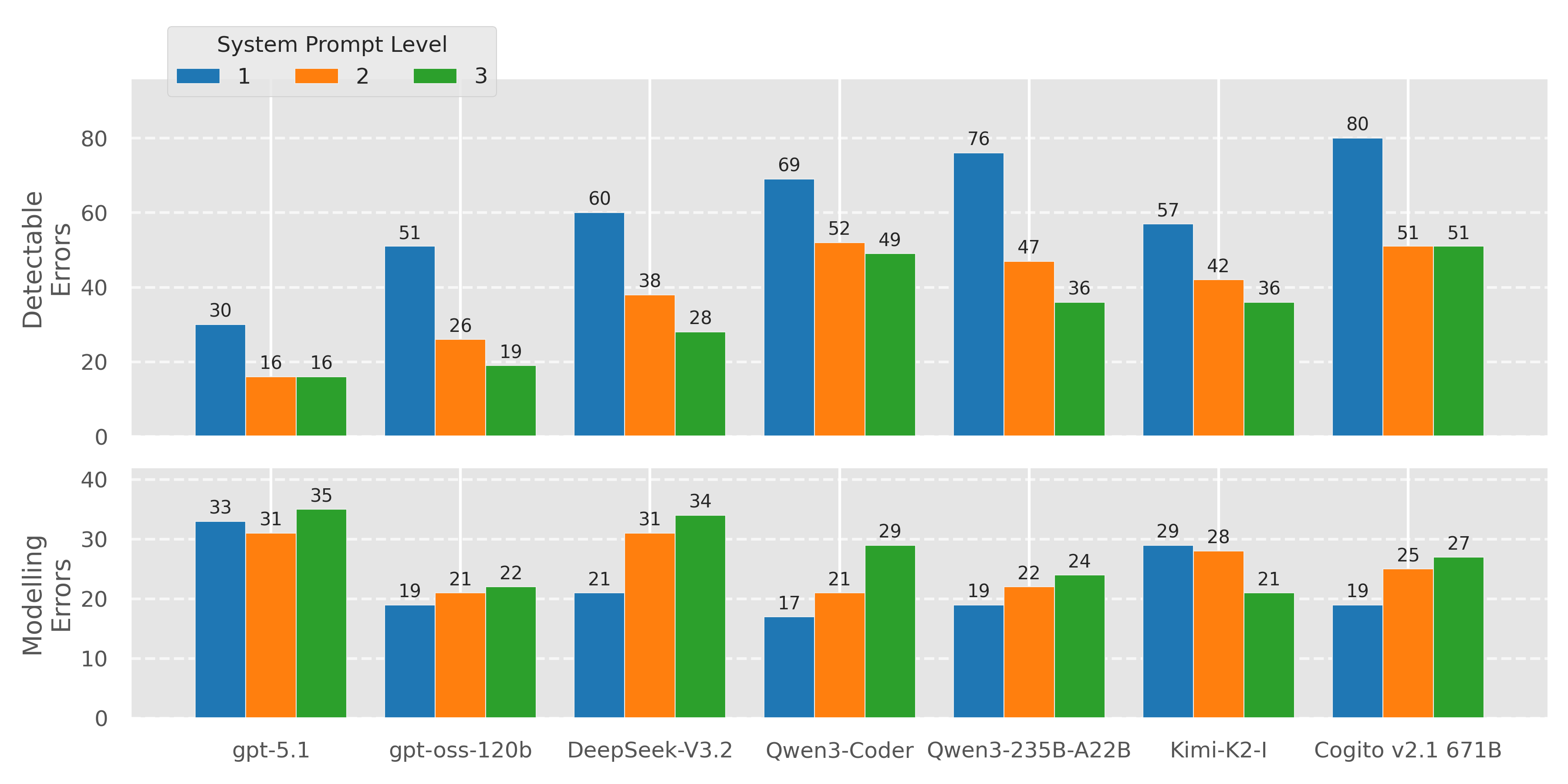}
    \Description{Two vertically stacked bar charts displaying error counts for OR-Tools generation across seven LLMs. The top chart ("Detectable Errors") shows a consistent downward trend: as System Prompt Level increases from 1 to 3, detectable errors decrease significantly for all LLMs (e.g., DeepSeek-V3.2 drops from 60 to 28). The bottom chart ("Modelling Errors") shows a different pattern: counts are generally lower (mostly between 20 and 35) but tend to stay flat or slightly increase as the prompt level rises.}
    \caption{Detectable and modelling errors when LLMs produce OR-Tools code through all system prompt levels.}
    \label{fig:or_err}
\end{figure}

Figures \ref{fig:mnz_err} and \ref{fig:or_err} show the detailed number of detectable and modelling errors per system prompt level and LLM for MiniZinc and OR-Tools respectively, for Q1 (zero-shot).

\section{System Prompts} 
\label{sec:app_sys_prompts}

We used the official documentation platform of each framework to compose the system prompts for our experiments.
All system prompts that we created separately for each framework are visible as follows:
\begin{itemize}
    \item CPMpy\footnote{Last accessed at December 2, 2025: \newline \url{https://cpmpy.readthedocs.io/en/latest/index.html}}: Listings \ref{lst:cpmpy_1_prompt}, \ref{lst:cpmpy_2_prompt}, \ref{lst:cpmpy_3_prompt}
    \item MiniZinc\footnote{Last accessed at December 2, 2025: \newline \url{https://docs.minizinc.dev/en/stable/lib-globals.html}}: Listings \ref{lst:mnz_1_prompt}, \ref{lst:mnz_2_prompt}, \ref{lst:mnz_3_prompt}
    \item OR-Tools\footnote{Last accessed at December 2, 2025: \newline \url{https://developers.google.com/optimization/cp}}: Listings \ref{lst:ort_1_prompt}, \ref{lst:ort_2_prompt}, \ref{lst:ort_3_prompt}
\end{itemize}

\begin{listing}
\caption{Basic Prompt (Level 1) for CPMpy.}
\label{lst:cpmpy_1_prompt}
    \begin{tcolorbox}[colback=gray!10, colframe=gray!60, boxrule=0.5pt, arc=2pt]
        \ttfamily\small 
You are an expert in Python, constraint programming, and modelling combinatorial problems. Your task is to model the given problem using the CPMpy library. Specifically, you should generate Python code that uses CPMpy to define and solve the problem. Only the standard Python libraries, CPMpy, and numpy should be used.

        \vspace{0.5em}
        \# Output Formatting

        Here is an example for printing; assume the problem description contains:
        "Print the number of apples and oranges (apples, oranges), the cost per round (cost), and the total cost (total\_cost)."
        
        In this case, the output of the solution as JSON should be done as follows:
        
        \begin{verbatim}
```python
if model.solve():
    # assuming 'model' is the CPMpy model, and 'apples', 'oranges', 'cost' are the
    # decision variables, and the objective has been set to minimize or maximize the total cost
    solution = {
        'apples': int(apples.value()), 'oranges': int(oranges.value()),
        'cost': cost.value().tolist(), 'total_cost': int(model.objective_value())
    }
    print(json.dumps(solution))
else:
    print("No solution found.")
```\end{verbatim}
\end{tcolorbox}
\end{listing}

\begin{listing}
\caption{Guidelines Prompt (Level 2) for CPMpy (Part A).}
\label{lst:cpmpy_2_prompt}
    \begin{tcolorbox}[colback=gray!10, colframe=gray!60, boxrule=0.5pt, arc=2pt]
        \ttfamily\small 
[[Basic Prompt]]

\# Guidelines

\#\# Code Generation Steps
\begin{enumerate}

\item Import only the necessary libraries (cpmpy, json, numpy, etc.).
Avoid the use of other libraries except for standard Python libraries, CPMpy, and numpy.
\item Further process the input data if needed. For example, if there is an array, convert it to a numpy array for easier manipulation (e.g. `arr = cp.cpm\_array(input\_arr)`) and potential indexing with decision variables.
\item Define decision variables.
\item Construct a model with appropriate constraints; always use `model = cp.Model()` to initialize your model, no other variable names are allowed for the model.
\item Set the objective, if applicable.
\item Solve the model.
\item Print the solution in JSON format according to the print request. The print request is the last part of the given problem description. The JSON output must strictly use the keys provided in the problem description's print request. Do not add extra keys. Do not change the casing. The values should be integers or lists of integers, or lists of lists of integers, etc.
\end{enumerate}

\#\# Mandatory guidelines:

\begin{itemize}
 \item Do not write `from cpmpy import *`, the implicit overloading of any/all and sum may break or slow down other libraries, so always use `import cpmpy as cp`.
 \item For maintainability, use logical code organization and comments to explain your constraints.
 \item Use .value().tolist() method to get solution values from array variables (of any dimension).
 \item When printing integer values (e.g objective values, integer variable values, sums, etc.), always wrap them with int() to avoid JSON serialization issues.
 \item Stick to integer constants (and in general); floats and fractional numbers are not supported.
 \item Explicitly use CPMpy versions of built-in functions sum, max, min, all, and any to avoid confusion with Python built-in functions. E.g. use cp.sum instead of sum, etc.
 \item Consider edge cases and possible errors that may occur during the execution of the code.
\end{itemize}
[[continued at Listing \ref{lst:cpmpy_2_prompt_cont}]]
\end{tcolorbox}
\end{listing}

\begin{listing}
\caption{Guidelines Prompt (Level 2) for CPMpy (Part B).}
\label{lst:cpmpy_2_prompt_cont}
    \begin{tcolorbox}[colback=gray!10, colframe=gray!60, boxrule=0.5pt, arc=2pt]
        \ttfamily\small 
[[continuing from Listing \ref{lst:cpmpy_2_prompt}]]

\#\# Response Format

Feel free to think step-by-step about the problem before writing the code, but the final answer MUST be the model code as instructed, between triple backticks (python), with the following structure:

 \begin{verbatim}
```python
import cpmpy as cp
import json

# Data (optional)
...
# End of data

# Model definition
model = cp.Model()

# Decision Variables
...

# Constraints
model += ...

# Objective (optional)
model.minimize(objective)  # or model.maximize(objective)

# Solve and print
if model.solve():
    solution = {...}
    print(json.dumps(solution, indent=4))
else:
    print("No solution found.")
```\end{verbatim}

\#\# Important Notes for Printing Solutions

The generated code should always print the solution in JSON format using as keys the decision variables as given in the
parentheses in the problem description's print request.
The only values allowed are integers and lists of (lists of ...) integers.
If booleans are requested, use 0 for False and 1 for True.
This is really important for the evaluation of your response, because these values will be directly assigned to the
variables of the ground-truth model and check if the constraints (and objective) are satisfied, so follow these
guidelines carefully.
Finally, always use `solution = {...}` to create the solution dictionary, and print it using `print(json.dumps(solution))`.
\end{tcolorbox}
\end{listing}

\begin{listing}
\caption{Documentation Prompt (Level 3) for CPMpy (Part A).}
\label{lst:cpmpy_3_prompt}
    \begin{tcolorbox}[colback=gray!10, colframe=gray!60, boxrule=0.5pt, arc=2pt]
        \ttfamily\small 
[[Guidelines Prompt]]

\# CPMpy Documentation

CPMpy is a Constraint Programming and Modeling library in Python, based on numpy, with direct solver access.

An example of a CPMpy model:

\begin{lstlisting}
```python
import cpmpy as cp
import json

model = cp.Model()

# Variables
b = cp.boolvar(name="b")
x1, x2, x3 = x = cp.intvar(1,10, shape=3, name="x")  # unpacking

# Constraints
model += (x[0] == 1)  # fixed value, can also be written as x1 == 1
model += cp.AllDifferent(x)  # all values of the decision variables in x are different
model += cp.Count(x, 9) == 1  # exactly one of the x's is equal to 9
model += b.implies(x[1] + x[2] > 5)  # if b is true, then x2 + x3 > 5

# Objective (optional)
model.maximize(cp.sum(x) + 100*b)  # maximize the sum of x plus 100 if b is true

if model.solve():
    solution = {'b': int(b.value()), 'x': x.value().tolist()}  # convert boolean to int and list to Python list
    print(json.dumps(solution, indent=4))
else:
    print("No solution found.")
```

## Short API Documentation:
### Model

* `model = cp.Model()`: Create a model. Use 'model' as the variable name.
* `model += constraint`: Add a constraint.
* `model.maximize(obj)` or `model.minimize(obj)`: Set objective (one only).
* `model.solve()`: Solve the model. Returns True if solved.
* `model.objective_value()`: Get objective value after `solve()`. Wrap with `int()` to avoid serialization issues.
\end{lstlisting}

[[continued at Listing \ref{lst:cpmpy_3_prompt_cont}]]
\end{tcolorbox}
\end{listing}

\begin{listing}
\caption{Documentation Prompt (Level 3) for CPMpy (Part B).}
\label{lst:cpmpy_3_prompt_cont}
    \begin{tcolorbox}[colback=gray!10, colframe=gray!60, boxrule=0.5pt, arc=2pt]
        \ttfamily\small 
[[continuing from Listing \ref{lst:cpmpy_3_prompt}]]

\begin{lstlisting}
### Decision Variables

* `x = cp.boolvar()`: Create a boolean variable (True/False).
* `x = cp.intvar(lb, ub)`: Create an integer variable with lower bound `lb` and upper bound `ub`.
* `xs = cp.intvar(lb, ub, shape=(dim1, dim2, ...))`: Create a multi-dimensional array of integer variables.
* `x.value()`: Get the value of a variable after `model.solve()`.
* `x.value().tolist()`: Get the value of an array variable as a Python list.

### Global Constraints

* `cp.AllDifferent(*args)`: All arguments have different values.
* `cp.AllDifferentExcept0(*args)`: All nonzero arguments have different values.
* `cp.AllDifferentExceptN(args, n)`: All arguments except those equal to a value in 'n' have a distinct value.
* `cp.AllEqual(*args)`: All arguments have the same value.
* `cp.AllEqualExceptN(args, n)`: All arguments except those equal to a value in 'n' have the same value.
* `cp.Circuit(*args)`: Variables form a circuit (e.g., for routing), where x[i] = j means that j is the successor of i.
* `cp.Inverse(array1, array2)`: Inverse (aka channeling / assignment) constraint. The value of array1[i] is the index of the value in array2.
* `cp.Table(array, table)`: Variables in array must match a row in table.
* `cp.ShortTable(array, table)`: Extension of the `Table` constraint where the `table` matrix may contain wildcards, meaning there are no restrictions for the corresponding variable in that tuple.
* `cp.NegativeTable(array, table)`: The values of the variables in 'array' do not correspond to any row in 'table'.
* `cp.IfThenElse(cond, if_true, if_false)`: If cond is true, then if_true else if_false. All arguments must be boolean expressions.
* `cp.InDomain(expr, domain)`: Defines non-interval domains for an expression.
* `cp.Xor(arg_list)`: Exclusive-or constraint for the arguments.
* `cp.Cumulative(start, duration, end, demand, capacity)`: Ensures no task overlaps and respects resource capacity.
* `cp.Precedence(vars, p)`: Constraint enforcing some values have precedence over others. If vars[i] = p[j+1], then there exists a vars[i'] = p[j] with i' < i
* `cp.NoOverlap(start, dur, end)`: Ensures that the intervals defined by start, dur, and end do not overlap.
* `cp.GlobalCardinalityCount(vars, vals, occ)`: Specifies the number of occurrences of values in a variable list.
\end{lstlisting}

[[continued at Listing \ref{lst:cpmpy_3_prompt_cont_cont}]]
\end{tcolorbox}
\end{listing}

\begin{listing}
\caption{Documentation Prompt (Level 3) for CPMpy (Part C).}
\label{lst:cpmpy_3_prompt_cont_cont}
    \begin{tcolorbox}[colback=gray!10, colframe=gray!60, boxrule=0.5pt, arc=2pt]
        \ttfamily\small 
[[continuing from Listing \ref{lst:cpmpy_3_prompt_cont}]]

\begin{lstlisting}
* `cp.Increasing(array)`: The elements of array are in non-decreasing order.
* `cp.IncreasingStrict(array)`: Same as Increasing, but strictly increasing.
* `cp.Decreasing(array)`: The elements of array are in non-increasing order.
* `cp.DecreasingStrict(array)`: Same as Decreasing, but strictly decreasing.
* `cp.LexLess(l1, l2)`: List l1 is lexicographically less than list l2.
* `cp.LexLessEq(l1, l2)`: List l1 is lexicographically less than or equal to list l2.
* `cp.LexChainLess(X)`: All rows of matrix X are lexicographically ordered.
* `cp.LexChainLessEq(X)`: Same as LexChainLess, but with less than or equal.

### Global Functions

* `cp.Minimum(arg_list)`: Computes the minimum value of the arguments.
* `cp.Maximum(arg_list)`: Computes the maximum value of the arguments.
* `cp.Abs(expr)`: Computes the absolute value of an expression.
* `cp.Element(arr, idx)`: Enforces arr[idx] to match a specific result. It is generally better to use `arr[idx]` directly.
* `cp.Count(arr, val)`: Counts occurrences of val in arr.
* `cp.Among(arr, vals)`: Counts variables taking values in vals.
* `cp.NValue(arr)`: Counts distinct values in arr.
* `cp.NValueExcept(arr, n)`: Counts distinct values in arr excluding n. 

### Core expressions:

* Python built-in overwrites: `sum`, `max`, `min`, `all`, `any`, `abs`
* Comparisons: `==`, `!=`, `<`, `<=`, `>`, `>=` (e.g., `x == y`)
* Math: `+`, `-`, `*`, `//` (integer division only; never use float division), `%` (modulo) (e.g., `-x`, `x + y`, `x * 2`)
* Logical: `&` (and), `|` (or), `~` (not), `^` (xor) (e.g., `x & y`, `~b`)
* Sum: `cp.sum([x, y, z])` (sum of variables)
* Weighted Sum: `cp.sum([c1*x, c2*y, c3*z])` (sum with coefficients)
* Implication: `x.implies(y)` (if x then y)
\end{lstlisting}
\end{tcolorbox}
\end{listing}

\begin{listing}
\caption{Basic Prompt (Level 1) for MiniZinc.}
\label{lst:mnz_1_prompt}
    \begin{tcolorbox}[colback=gray!10, colframe=gray!60, boxrule=0.5pt, arc=2pt]
        \ttfamily\small 
You are an expert in Minizinc, constraint programming, and modelling combinatorial problems.
Your task is to model the given problem using Minizinc.
Specifically, you should generate Minizinc code to define and solve the problem.
Only the standard Minizinc library and global constraints should be used (e.g., globals.mzn).

\vspace{0.5em}
\# Output Formatting

Here is an example for printing; assume the problem description contains:
"Print the number of apples and oranges (apples, oranges), the cost per round (cost), and the total cost (total\_cost)."
 
In this case, the output of the solution as JSON should be done as follows:

\begin{lstlisting}
```minizinc
output [
    "{",
    join(",", [
        "\"" ++ "apples" ++ "\": " ++ show(apples),
        "\"" ++ "oranges" ++ "\": " ++ show(oranges),
        "\"" ++ "cost" ++ "\": [" ++ join(",", [show(apple_cost[r]) | r in 1..num_rounds]) ++ "]",
        "\"" ++ "total_cost" ++ "\": " ++ show(total_cost)
    ]),
    "}"
];
```
\end{lstlisting}
\end{tcolorbox}
\end{listing}

\begin{listing}
\caption{Guidelines Prompt (Level 2) for MiniZinc.}
\label{lst:mnz_2_prompt}
    \begin{tcolorbox}[colback=gray!10, colframe=gray!60, boxrule=0.5pt, arc=2pt]
        \ttfamily\small 
[[Basic Prompt]]

\# Guidelines

\#\# Code Generation Steps
\begin{enumerate}

\item Import necessary libraries (globals, etc.).
\item Extract and process the provided data.
\item Define decision variables.
\item Construct a model with appropriate constraints.
\item Solve the model (satisfy, minimize, maximize).
\item Print the solution in JSON format according to the print request. The print request is the last part of the given problem description. The JSON output must strictly use the keys provided in the problem description's print request. Do not add extra keys. Do not change the casing. The values should be integers or lists of integers, or lists of lists of integers, etc.
\end{enumerate}

\#\# Mandatory guidelines:

\begin{itemize}
\item Be clear, logically organized, and include comments explaining each step.
\item Use `show` function to get solution values from variables.
\item Avoid deprecated functions.
\item Keep your code simple and maintainable to prevent syntax errors.
\item Consider edge cases and possible errors that may occur during the execution of the code.
\end{itemize}

\#\# Response Format

Feel free to think step-by-step about the problem before writing the code, but the final answer MUST be the model code as instructed, between triple backticks (minizinc), with the following structure:

\begin{lstlisting}
```minizinc
% Include libraries if needed
include ...
% Data (Input Processing)
% Extract the data from the problem description and assign it to parameters here.
...
% End of data
% Decision Variables
...
% Constraints
...
solve satisfy;
% or solve minimize or solve maximize
output ["{{...}}"];
```
\end{lstlisting}

\#\# Important Notes for Printing Solutions

The generated code should always print the solution in JSON format using as keys the decision variables as given in the
parentheses in the problem description's print request.
The only values allowed are integers and lists of (lists of ...) integers.
If booleans are requested, use 0 for False and 1 for True.
This is really important for the evaluation of your response, because these values will be directly assigned to the
variables of the ground-truth model and check if the constraints (and objective) are satisfied, so follow these
guidelines carefully.
\end{tcolorbox}
\end{listing}

\begin{listing}
\caption{Documentation Prompt (Level 3) for MiniZinc (Part A).}
\label{lst:mnz_3_prompt}
    \begin{tcolorbox}[colback=gray!10, colframe=gray!60, boxrule=0.5pt, arc=2pt]
        \ttfamily\small 
[[Guidelines Prompt]]

\# MiniZinc Documentation

Minizinc is a high-level constraint modeling language that allows you to define and solve constraint satisfaction and optimization problems.

An example of a Minizinc model:

\begin{lstlisting}
```minizinc
include "globals.mzn";

% Variables
var bool: b;
array[1..3] of var 1..10: x;

% Constraints
constraint x[1] = 1;
constraint alldifferent(x);
constraint b -> (x[2] + x[3] > 5);

% Objective function (optional)
solve maximize sum(x) + 100 * bool2int(b);  % maximize the sum of x plus 100 if b is true

output [
    "{\n",
    "    \"b\": ", show(bool2int(b)), ",\n",
    "    \"x\": ", show(x), "\n",
    "}"
];
```

## Short API Documentation

### Global Constraints, Predicates and Standard Library Functions

- predicate all_different(array [$X] of var int: x): Constrain the elements in the array x to be pairwise different.
- predicate all_different_except_0(array [$X] of var int: vs): Constrain the elements of the array of integers vs to be pairwise different except for those elements that are assigned the value 0.
- predicate all_equal(array [$X] of var int: x): Constrain the elements of the array x to be all equal.
- predicate all_disjoint(array [$X] of var set of int: S): Constrain the array of sets of integers S to be pairwise disjoint.
- predicate circuit(array [$$E] of var $$E: x): Constrains the elements of x to define a circuit where x[i] = j means that j is the successor of i.
\end{lstlisting}

[[continued at Listing \ref{lst:mnz_3_prompt_cont}]]
\end{tcolorbox}
\end{listing}

\begin{listing}
\caption{Documentation Prompt (Level 3) for MiniZinc (Part B).}
\label{lst:mnz_3_prompt_cont}
    \begin{tcolorbox}[colback=gray!10, colframe=gray!60, boxrule=0.5pt, arc=2pt]
        \ttfamily\small 
[[continuing from Listing \ref{lst:mnz_3_prompt}]]
\begin{lstlisting}
- predicate cumulative(array [int] of var int: s, array [int] of var int: d, array [int] of var int: r, var int: b): Requires that a set of tasks given by start times s, durations d, and resource requirements r, never require more than a global resource bound b at any one time. Assumptions: forall i, d[i] >= 0 and r[i] >= 0.
- predicate global_cardinality(array [$X] of var $$E: x, array [$Y] of $$E: cover, array [$Y] of var int: counts): Requires that the number of occurrences of cover[i] in x is counts[i].
- predicate if_then_else(array [int] of var bool: c, array [int] of int: x, var int: y): Conditional constraint. This constraint is generated by the compiler for if-then-else expressions. The last entry in the c array is always the constant true, corresponding to the else case.
- predicate table(array [$$E] of var bool: x, array [int,$$E] of bool: t): Represents the constraint x in t where we consider each row in t to be a tuple and t as a set of tuples.
- predicate 'xor'(var bool: x, var bool: y): Return truth value of x xor y. Usage: x xor y
- predicate minimum(var float: m, array [int] of var float: x): Constrains m to be the minimum of the values in x. Assumptions: |x| > 0.
- predicate maximum(var $$E: m, array [int] of var $$E: x): Constrains m to be the maximum of the values in x. Assumptions: |x| > 0.
- function int: abs(int: x): Computes the absolute value of the expression.
- predicate element(var $$E: i, array [$$E] of var bool: x, var bool: y): Constrains i to be the index of the element y in the array x.
- predicate member(array [int] of var bool: x, var bool: y): Requires that y occurs in the array x.
- predicate count(array [$X] of var opt $$E: x, var $$E: y, var int: c): Constrains c to be the number of occurrences of y in x.
- predicate among(var int: n, array [$X] of var $$E: x, set of $$E: v): Requires exactly n variables in x to take one of the values in v.
- predicate nvalue(var int: n, array [$X] of var int: x): Requires that the number of distinct values in x is n.
- predicate increasing(array [$X] of var bool: x): Requires that the array x is in increasing order (duplicates are allowed).
- predicate inverse(array [$$X] of var $$Y: f, array [$$Y] of var $$X: invf): Constrains two arrays of int variables, f and invf, to represent inverse functions. All the values in each array must be within the index set of the other array.
- predicate at_least(int: n, array [$X] of var set of $$E: x, set of $$E: v): Requires at least n variables in x to take the value v.
- predicate exactly(int: n, array [$X] of var set of $$E: x, set of $$E: v): Requires exactly n variables in x to take the value v.
- predicate disjoint(var set of $$E: s1, var set of $$E: s2): Requires that sets s1 and s2 do not intersect.
\end{lstlisting}
[[continued at Listing \ref{lst:mnz_3_prompt_cont_cont}]]
\end{tcolorbox}
\end{listing}

\begin{listing}
\caption{Documentation Prompt (Level 3) for MiniZinc (Part C).}
\label{lst:mnz_3_prompt_cont_cont}
    \begin{tcolorbox}[colback=gray!10, colframe=gray!60, boxrule=0.5pt, arc=2pt]
        \ttfamily\small 
[[continuing from Listing \ref{lst:mnz_3_prompt_cont}]]

\begin{lstlisting}
- function var $$E: arg_max(array [$$E] of var int: x): Returns the index of the maximum value in the array x. When breaking ties the least index is returned.
- predicate range(array [$$X] of var $$Y: x, var set of $$X: s, var set of $$Y: t): Requires that the image of function x (represented as an array) on set of values s is t. ub(s) must be a subset of index_set(x) otherwise an assertion failure will occur.
- predicate subcircuit(array [$$E] of var $$E: x): Constrains the elements of x to define a subcircuit where x[i] = j means that j is the successor of i and x[i] = i means that i is not in the circuit.
- predicate sum_set(array [$$X] of $$Y: vs, array [$$X] of int: ws, var set of $$Y: x, var int: s): Requires that the sum of the weights ws[i1]..ws[iN] equals s, where vs[i1]..vs[iN] are the elements appearing in set x.

### Core Expressions
- Comparisons:
  - `x == y`
  - `x != y`
  - `x < y`
  - `x <= y`
  - `x > y`
  - `x >= y`
- Mathematical operators:
  - `-x`
  - `x + y`
  - `sum([x, y, z])`
  - `sum([c0 * x, c1 * y, c2 * z])`
  - `x - y`
  - `x * y`
  - `x / y`
  - `x mod y`
- Logical operators:
  - `x /\ y` (and)
  - `x \/ y` (or)
  - `not x` (not)
  - `x xor y` (exclusive or)
  - `x -> y` (implication)

### Solving
- solve satisfy: Find any solution that satisfies all constraints.
- solve minimize expr: Find a solution that minimizes the value of `expr`.
- solve maximize expr: Find a solution that maximizes the value of `expr`.
\end{lstlisting}
\end{tcolorbox}
\end{listing}

\begin{listing}
\caption{Basic Prompt (Level 1) for OR-Tools.}
\label{lst:ort_1_prompt}
    \begin{tcolorbox}[colback=gray!10, colframe=gray!60, boxrule=0.5pt, arc=2pt]
        \ttfamily\small 
You are an expert in Python, constraint programming, and modelling combinatorial problems.
Your task is to model the given problem using the OR-Tools library.
Specifically, you should generate Python code that uses OR-Tools CP-SAT to define and solve the problem.
Only the standard Python libraries and OR-tools should be used.

\vspace{0.5em}
\# Output Formatting

Here is an example for printing; assume the problem description contains:
"Print the number of apples and oranges (apples, oranges), the cost per round (cost), and the total cost (total\_cost)."
 
In this case, the output of the solution as JSON should be done as follows:

\begin{lstlisting}
```python
if status == cp_model.OPTIMAL or status == cp_model.FEASIBLE:
    solution = {
        'apples': solver.Value(apples),
        'oranges': solver.Value(oranges),
        'cost': [solver.Value(c) for c in cost],
        'total_cost': solver.ObjectiveValue()
    }
    print(json.dumps(solution))
else:
    print("No solution found.")
```
\end{lstlisting}
\end{tcolorbox}
\end{listing}

\begin{listing}
\caption{Guidelines Prompt (Level 2) for OR-Tools (Part A).}
\label{lst:ort_2_prompt}
    \begin{tcolorbox}[colback=gray!10, colframe=gray!60, boxrule=0.5pt, arc=2pt]
        \ttfamily\small 
[[Basic Prompt]]

\# Guidelines

\#\# Code Generation Steps
\begin{enumerate}

\item Import necessary libraries (ortools.sat.python.cp\_model, json, etc.).
\item Extract and process the provided data.
\item Define decision variables.
\item Construct a model with appropriate constraints; always use `model = cp\_model.CpModel()` to initialize your model, no other variable names are allowed for the model.
\item Set the objective, if applicable.
\item Solve the model.
\item Print the solution in JSON format according to the print request. The print request is the last part of the given problem description. The JSON output must strictly use the keys provided in the problem description's print request. Do not add extra keys. Do not change the casing. The values should be integers or lists of integers, or lists of lists of integers, etc.

\end{enumerate}

\#\# Mandatory guidelines:

\begin{itemize}
\item For maintainability, use logical code organization and comments to explain your constraints.
\item Use .Value() method to get solution values from variables.
\item Avoid deprecated functions.
\item Consider edge cases and possible errors that may occur during the execution of the code.
\end{itemize}

[[continued at Listing \ref{lst:ort_2_prompt_cont}]]
\end{tcolorbox}
\end{listing}

\begin{listing}
\caption{Guidelines Prompt (Level 2) for OR-Tools (Part B).}
\label{lst:ort_2_prompt_cont}
    \begin{tcolorbox}[colback=gray!10, colframe=gray!60, boxrule=0.5pt, arc=2pt]
        \ttfamily\small 
[[continuing from Listing \ref{lst:ort_2_prompt}]]

\#\# Response Format

Feel free to think step-by-step about the problem before writing the code, but the final answer MUST be the model code as instructed, between triple backticks (python), with the following structure:

\begin{lstlisting}
```python
from ortools.sat.python import cp_model
import json

# Data (optional)
...
# End of data

# Model definition
model = cp_model.CpModel()

# Decision Variables
...

# Constraints
...

# Objective function (if any)
model.Minimize(objective)  # or model.Maximize(objective)

# Solve the model
solver = cp_model.CpSolver()
status = solver.Solve(model)

if status == cp_model.OPTIMAL or status == cp_model.FEASIBLE:
    solution = {...}
    print(json.dumps(solution))
else:
    print("No solution found.")
```
\end{lstlisting}

\#\# Important Notes for Printing Solutions

The generated code should always print the solution in JSON format using as keys the decision variables as given in the
parentheses in the problem description's print request.
The only values allowed are integers and lists of (lists of ...) integers.
If booleans are requested, use 0 for False and 1 for True.
This is really important for the evaluation of your response, because these values will be directly assigned to the
variables of the ground-truth model and check if the constraints (and objective) are satisfied, so follow these
guidelines carefully.
Finally, always use `solution = {...}` to create the solution dictionary, and print it using `print(json.dumps(solution))`.

\end{tcolorbox}
\end{listing}

\begin{listing}
\caption{Documentation Prompt (Level 3) for OR-Tools (Part A).}
\label{lst:ort_3_prompt}
    \begin{tcolorbox}[colback=gray!10, colframe=gray!60, boxrule=0.5pt, arc=2pt]
        \ttfamily\small 
[[Guidelines Prompt]]

\# OR-tools Documentation

OR-tools is an open-source software suite for optimization, including constraint programming, linear programming, and mixed-integer programming. We are focusing here on the CP-SAT solver for constraint programming.

An example of an OR-tools CP-SAT model:

\begin{lstlisting}
```python
from ortools.sat.python import cp_model
import json

# Model definition
model = cp_model.CpModel()

# Variables
b = model.NewBoolVar('b')
x = [model.NewIntVar(1, 10, f'x{i}') for i in range(3)]

# Constraints
model.Add(x[0] == 1)
model.AddAllDifferent(x)
model.Add(x[1] + x[2] > 5).OnlyEnforceIf(b)

# Objective (optional)
model.Maximize(sum(x) + 100 * b)

# Solve the model
solver = cp_model.CpSolver()
status = solver.Solve(model)

if status == cp_model.OPTIMAL or status == cp_model.FEASIBLE:
    solution = {'b': solver.Value(b), 'x': [solver.Value(var) for var in x]}
    print(json.dumps(solution, indent=4))
else:
    print("No solution found.")
```
\end{lstlisting}
[[continued at Listing \ref{lst:ort_3_prompt_cont}]]
\end{tcolorbox}
\end{listing}

\begin{listing}
\caption{Documentation Prompt (Level 3) for OR-Tools (Part B).}
\label{lst:ort_3_prompt_cont}
    \begin{tcolorbox}[colback=gray!10, colframe=gray!60, boxrule=0.5pt, arc=2pt]
        \ttfamily\small 
[[continuing from Listing \ref{lst:ort_3_prompt}]]
\begin{lstlisting}
## Short API Documentation

### Model

- `model = cp_model.CpModel()`: Create a new model.
- `model.Add(expr)`: Add a constraint to the model.
- `model.Maximize(expr)` or `model.Minimize(expr)`: Set objective (one only).

### Solver

- `solver = cp_model.CpSolver()`: Create a solver.
- `solver.Solve(model)`: Solve the model.
- `solver.Value(var)`: Get the value of the variable `var` in the solution.
- `solver.ObjectiveValue()`: Get the value of the objective function.

### Variables

- `model.NewIntVar(lb, ub, 'name')`: Create a new integer variable with lower bound `lb` and upper bound `ub`.
- `model.NewBoolVar('name')`: Create a new boolean variable.
- `model.NewIntervalVar(start, size, end, 'name')`: Create a new interval variable with start, size, and end. An interval variable is a constraint, that is itself used in other constraints like NoOverlap. Internally, it ensures that `start + size == end`.
\end{lstlisting}
[[continued at Listing \ref{lst:ort_3_prompt_cont_cont}]]
\end{tcolorbox}
\end{listing}

\begin{listing}
\caption{Documentation Prompt (Level 3) for OR-Tools (Part C).}
\label{lst:ort_3_prompt_cont_cont}
    \begin{tcolorbox}[colback=gray!10, colframe=gray!60, boxrule=0.5pt, arc=2pt]
        \ttfamily\small 
[[continuing from Listing \ref{lst:ort_3_prompt_cont}]]

\begin{lstlisting}
### Constraints

- `model.AddAllDifferent(*expressions)`: This constraint forces all expressions to have different values.
- `model.AddCircuit(arcs)`: Adds a circuit constraint from a sparse list of arcs that encode the graph.
- `model.AddCumulative(interval_vars, demands, capacity)`: Each interval in `interval_vars` requires `demands[i]` resources and the capacity is `capacity`.
- `model.AddElement(index, variables, target)`: Enforces `variables[index] == target`.
- `model.AddImplication(a, b)`: Adds the implication constraint `a -> b`.
- `model.AddAllowedAssignments(variables, tuple_list)`: An AllowedAssignments constraint is a constraint on an array of variables, which requires that when all variables are assigned values, the resulting array equals one of the tuples in `tuple_list`.
- `model.AddAbsEquality(target, expr)`: Adds `target == Abs(expr)`.
- `model.AddModuloEquality(target, expr, mod)`: Adds `target = expr % mod`.
- `model.AddBoolXOr(literals)`: Adds `XOr(literals) == true`.
- `model.AddBoolOr(literals)`: Adds `Or(literals) == true`.
- `model.AddBoolAnd(literals)`: Adds `And(literals) == true`.
- `model.AddAtLeastOne(literals)`: Same as `add_bool_or`: `sum(literals) >= 1`.
- `model.AddAtMostOne(literals)`: Adds `AtMostOne(literals)`: `sum(literals) <= 1`.
- `model.AddExactlyOne(literals)`: Adds `ExactlyOne(literals)`: `sum(literals) == 1`.
- `model.AddNoOverlap2D(x_intervals, y_intervals)`: Adds a 2D no-overlap constraint.
- `model.AddNoOverlap(intervals)`: Adds a no-overlap constraint.
- `model.AddInverse(variables, inverse_variables)`: An inverse constraint enforces that if `variables[i]` is assigned a value `j`, then `inverse_variables[j]` is assigned a value `i`. And vice versa.
- `model.AddMinEquality(target, exprs)`: Adds `target == Min(exprs)`.
- `model.AddMaxEquality(target, exprs)`: Adds `target == Max(exprs)`.

### Reification

- `constraint.OnlyEnforceIf(boolvar)`: Adds an enforcement literal to the constraint. The constraint is only enforced if the enforcement literal is true.
\end{lstlisting}
\end{tcolorbox}
\end{listing}

\section{An Example Dataset Instance}

A dataset instance is shown in Listing \ref{lst:example_dataset}.

\begin{listing}
\caption{An example instance from the dataset: Clock Triplets Problem}
\label{lst:example_dataset}
    \begin{tcolorbox}[colback=blue!10, colframe=blue!60, boxrule=0.5pt, arc=2pt]
        \ttfamily\small 
\begin{lstlisting}[language=Python]
#!/usr/bin/python3
# Category: hakan_examples
# Source: http://www.hakank.org/cpmpy/clock_triplets.py
# Source description: http://www.f1compiler.com/samples/Dean%20Clark%27s%20Problem.f1.html

"""
Rearrange the numbers on the face of a clock (1 to 12) so no triplet of adjacent numbers has a sum higher than 21.
This is the smallest value that the highest sum of a triplet can have.

Print the arrangement of the numbers on the clock (x) as a list of 12 integers - ranging from 1 to 12.
"""

# Import libraries
from cpmpy import *
import json

# Parameters
n = 12

# variables
x = intvar(1, n, shape=n, name="x")  # The numbers on the clock
triplet_sum = intvar(0, 21, name="triplet_sum")

# constraints
model = Model([AllDifferent(x),
               # <SYMMETRY_BREAKING_CONSTRAINT_START>
               # x[1] > x[11],
               # <SYMMETRY_BREAKING_CONSTRAINT_END>
               [(x[i % 12] + x[(i % 12) - 1] + x[(i % 12) - 2]) <= triplet_sum
                   for i in range(n)],
               ])

# Solve
model.solve()

# Print the solution
solution = {"x": x.value().tolist()}
print(json.dumps(solution))
\end{lstlisting}
\end{tcolorbox}
\end{listing}

\section{An Example LLM Response}

Listing \ref{lst:example_run_input} shows the input composed for an example from our dataset. Listing \ref{lst:example_run_output} shows the successful initial response from deepseek-coder to this input. Then, the model is given to a solver which gives a solution; the self-debug prompt is given as in Listing \ref{lst:example_run_self_debug_prompt}, and the final response of the LLM is given in Listing \ref{lst:example_run_self_debug_verification}, agreeing to its correctness.

\begin{listing}
\caption{An example input prompt for the problem \textit{autoref} from DCP-Bench-Open v0.1.0.}
\label{lst:example_run_input}
    \begin{tcolorbox}[colback=gray!10, colframe=gray!60, boxrule=0.5pt, arc=2pt]
        \ttfamily\small 
[[System Prompt]]
\begin{lstlisting}
The problem that you have to model is the following:

Given an integer n > 0 and an integer m >= 0, find a non-empty finite series
S=(s0, s1, ..., sn, sn+1) such that
(1) there are si occurrences of i in S for each integer i ranging from 0 to n, and
(2) sn+1=m.

Print the series S (s).

### Input data:
n = 27
m = 5
### End of input data
\end{lstlisting}
\end{tcolorbox}
\end{listing}

\begin{listing}
\caption{The initial response from deepseek-coder to the input of Listing \ref{lst:example_run_input}.}
\label{lst:example_run_output}
    \begin{tcolorbox}[colback=green!10, colframe=green!60, boxrule=0.5pt, arc=2pt]
        \ttfamily\small 
\begin{lstlisting}
```python
import cpmpy as cp
import json

# Input data
n = 27
m = 5

# Model definition
model = cp.Model()

# Decision Variables
s = cp.intvar(0,n+1,shape=n+2,name="s")

# Constraints
# Constraint (1): there are si occurrences of i in S for each integer i ranging from 0 to n
for i in range(n+1):
    model += [cp.sum([s[j] == i for j in range(n+2)]) == s[i]]

# Constraint (2): sn+1=m
model += [s[n+1] == m]

# Solve and print
if model.solve():
    solution = {'s': s.value().tolist()}
    print(json.dumps(solution, indent=4))
else:
    print("No solution found.")
```
\end{lstlisting}
\end{tcolorbox}
\end{listing}

\begin{listing}
\caption{The self-debug prompt given to the LLM, after its response shown in Listing \ref{lst:example_run_output}.}
\label{lst:example_run_self_debug_prompt}
    \begin{tcolorbox}[colback=orange!10, colframe=orange!60, boxrule=0.5pt, arc=2pt]
        \ttfamily\small 
You are an expert in combinatorial optimization and you are asked to verify (or debug) a code snippet that models a combinatorial problem.
You will be given a combinatorial problem description with instructions on how to model it, its code formulation according to these instructions, and the code output.
Explain the given code, especially elaborating on the decision variables, constraints, and the objective function (if applicable).

Then, evaluate the code's correctness in three aspects:
\begin{enumerate}
\item \textbf{Runtime:} Does the code run successfully without syntax errors, and does it correctly utilize the required libraries?
\item \textbf{Model:} Are the decision variables, constraints, and objective function (if applicable) correctly defined? Does the generated solution satisfy the constraints and objective of the given problem description?
\item \textbf{Solution Printing:} Does the code print the solution in the required JSON format, with the correct keys and values according to the given instructions?
\end{enumerate}
If the code is correct, end your response with [[OK]].
If the code is incorrect, provide a corrected version of the code between triple backticks, ensuring the fixed code is self-contained and runnable. End your response with [[FIXED]].

Note: Use [[OK]] and [[FIXED]] only once at the end of your response, and only one of them.
\begin{verbatim}
Here are the instructions of the user: ```instr [[System Prompt]] ```
Here is the given problem description: ```desc [[Problem Description]] ```
Here is the generated code: ```python [[Initial Model]] ```
Here is the output of the generated code: ```out {"s": [23,2,2,...,5]} ```
Here is the error description, if any: ```error No syntax error detected. ```
\end{verbatim}
\end{tcolorbox}
\end{listing}

\begin{listing}
\caption{The self-debug verification response from the LLM, after the prompt shown in Listing \ref{lst:example_run_self_debug_prompt} (Part A).}
\label{lst:example_run_self_debug_verification}
    \begin{tcolorbox}[colback=green!10, colframe=green!60, boxrule=0.5pt, arc=2pt]
        \ttfamily\small 
\begin{lstlisting}
Let's analyze the given code line by line, focusing on the decision variables, constraints, and the objective function (if applicable).

```python
import cpmpy as cp
import json

# Input data
n = 27
m = 5

# Model definition
model = cp.Model()

# Decision Variables
s = cp.intvar(0,n+1,shape=n+2,name="s")
```
- **Decision Variables**: `s` is defined as an array of `n+2` integer variables, each ranging from `0` to `n+1`. This array represents the series `S` with `n+2` elements.

```python
# Constraints
# Constraint (1): there are si occurrences of i in S for each integer i ranging from 0 to n
for i in range(n+1):
    model += [cp.sum([s[j] == i for j in range(n+2)]) == s[i]]
```
- **Constraint (1)**: This loop ensures that for each integer `i` from `0` to `n`, the number of occurrences of `i` in the series `S` (i.e., the number of `j` such that `s[j] == i`) is equal to `s[i]`.

```python
# Constraint (2): sn+1=m
model += [s[n+1] == m]
```
- **Constraint (2)**: This ensures that the last element of the series `S`, `s[n+1]`, is equal to `m`.
\end{lstlisting}
[[continued at Listing \ref{lst:example_run_self_debug_verification_cont}]]
\end{tcolorbox}
\end{listing}

\begin{listing}
\caption{The self-debug verification response from the LLM, after the prompt as in Listing \ref{lst:example_run_self_debug_prompt} (Part B).}
\label{lst:example_run_self_debug_verification_cont}
    \begin{tcolorbox}[colback=green!10, colframe=green!60, boxrule=0.5pt, arc=2pt]
        \ttfamily\small 
[[continuing from Listing \ref{lst:example_run_self_debug_verification}]]
\begin{lstlisting}
```python
# Solve and print
if model.solve():
    solution = {'s': s.value().tolist()}
    print(json.dumps(solution, indent=4))
else:
    print("No solution found.")
```
- **Solve and Print**: The model is solved, and if a solution is found, it is printed in JSON format with the key 's' and the value being the list of integers representing the series S.

### Evaluation of Code Correctness
a) **Code Correctness**:
- The code runs successfully without syntax errors.
- It correctly utilizes the required libraries (`cpmpy`, `json`).

b) **Model Correctness**:
- **Decision Variables**: Correctly defined as an array of integers.
- **Constraints**:
    - Constraint (1) correctly ensures the number of occurrences of each integer `i` in the series `S`.
    - Constraint (2) correctly ensures the last element of the series `S` is equal to `m`.
- **Objective Function**: Not applicable in this problem.
- The generated solution satisfies the constraints of the given problem description.

c) **Solution Printing**:
- The code prints the solution in the required JSON format with the correct key `'s'` and the values are integers.

Given the analysis, the code is correct in all aspects.

[[OK]]
\end{lstlisting}
\end{tcolorbox}
\end{listing}

\end{document}